# Four-hour thunderstorm nowcasting using a deep diffusion model for satellite data

## Author Information

Kuai Dai[1,3], Xutao Li[1]*, Junying Fang[2], Yunming Ye[1]*, Demin Yu[1], Hui Su[3], Di Xian[4], Danyu Qin[4], Jingsong Wang[4]*

## Affiliations

**[1]School of Computer Science and Technology, Harbin Institute of Technology, Shenzhen, China**

**[2]Institute of Tropical and Marine Meteorology, China Meteorological Administration, Guangzhou, China**

**[3]State Key Laboratory of Climate Resilience for Coastal Cities, Department of Civil and Environmental Engineering, Hong Kong University of Science and Technology, Hong Kong, China**

**[4]National Satellite Meteorological Center, China Meteorological Administration, Beijing, China**

## Corresponding authors

Correspondence to: Xutao Li, lixutao@hit.edu.cn; Yunming Ye, yeyunming@hit.edu.cn; Jingsong Wang, wangjs@cma.gov.cn

## Abstract

Convection (thunderstorm) develops rapidly within hours and is highly destructive, posing a significant challenge for nowcasting and resulting in substantial losses to infrastructure and society. After the emergence of artificial intelligence (AI)-based methods, convection nowcasting has experienced rapid advancements, with its performance surpassing that of physics-based numerical weather prediction and other conventional approaches. However, the lead time and coverage of it still leave much to be desired and hardly meet the needs of disaster emergency response. Here, we propose a **d**eep **d**iffusion **m**odel for **s**atellite data (DDMS) to establish an AI-based convection nowcasting system. Specifically, DDMS employs diffusion processes to effectively simulate complicated spatiotemporal evolution patterns of convective clouds, achieving more accurate forecasts of convective growth and dissipation over longer lead times. Additionally, it combines geostationary satellite brightness temperature data and domain knowledge from meteorological experts, thereby achieving planetary-scale forecast coverage. During long-term tests and objective validation based on the FengYun-4A satellite, our system achieves, for the first time, effective convection nowcasting up to 4 hours, with broad coverage (about 20,000,000 $km^2$), remarkable accuracy, and high resolution (15 minutes; 4 km). Its performance reaches a new height in convection nowcasting compared to the existing models. In terms of application, our system is highly transferable with the potential to collaborate with multiple satellites for global convection nowcasting. Furthermore, our results highlight the remarkable capabilities of diffusion models in convective clouds forecasting, as well as the significant value of geostationary satellite data when empowered by AI technologies.

## Introduction

Severe convective weather often leads to sudden meteorological disasters like hailstorms, thunderstorms, strong winds, and tornadoes[1-3]. These events have significant adverse effects on various aspects of daily life and economic activities such as transportation, agriculture, energy production, and societal management[4,5]. As a conventional approach for convective weather nowcasting and forecasting, the numerical weather prediction (NWP) solves the physical equations in the atmosphere to make predictions[6,7]. While traditional NWP has advantages in large-scale weather forecast from several hours to days ahead, it has historically struggled with localized nowcasting due to challenges in resolving rapidly evolving mesoscale or convective-scale processes[8,9]. However, advances such as the Warn-on-Forecast System (WoFS)[10,11] have demonstrated the potential of high-resolution modeling and rapid data assimilation to improve probabilistic convective-scale predictions. In parallel, advection-based methods such as STEPS[12] or PySTEPS[13] remain widely used alternatives, leveraging physical motion principles to extrapolate current radar or satellite observations into short-term forecasts. The advection-based approaches continue struggling to provide precise nowcasting outcomes due to two primary factors. First, the estimated advection tendency is unable to capture the complexity of atmospheric variability, such as cloud formation and dissipation. Second, these methods are limited in their ability to leverage extensive historical data for modeling nonlinear spatiotemporal dynamics, since they estimate the motion field based solely on the recent observations[14].

In recent years, artificial intelligence (AI) technologies largely promote the forward of weather nowcasting. With advanced neural network architectures[8,15,16], the AI-based nowcasting models learn spatiotemporal evolution patterns from historical meteorological data in an end-to-end optimization manner. Once trained, such models can efficiently make predictions with given observations. Specifically, the deep generative model of rainfall (DGMR)[8] introduces generative adversarial networks (GANs)[17] to improve the nowcasting results and achieves spatiotemporally consistent predictions with a lead time of 90 minutes. As a state-of-the-art precipitation nowcasting approach, NowcastNet[15] designs a special evolution network to learn physical knowledge to guide prediction and achieves 3-hour extreme precipitation prediction. Despite the progress delivered by such existing AI-based nowcasting approaches, they still fall short due to their limited lead time, spatial scale, and accuracy, especially in heavy rainfall closely related to convection.

As satellite remote sensing technologies advance rapidly, the latest generation of geostationary meteorological satellites now offers extensive atmospheric monitoring capabilities, enabling them to detect thunderstorm initiation earlier than radars[18-23]. By utilizing the infrared brightness temperature band, these satellites achieve high spatiotemporal resolution convective monitoring, which is quite close to that of ground-based radars. Specifically, lower brightness temperatures correspond to colder and higher cloud tops, which are widely recognized as a proxy for stronger convective activity. Furthermore, convection nowcasting with geostationary satellite has gained widespread adoption in meteorological departments globally[26-28]. Compared to nowcasting convective activities based on ground-based radar data (whether from single stations or networked), a prominent advantage of nowcasting convection using brightness temperatures from geostationary satellites is the ability to monitor large areas at a planetary scale, including continuous 24-hour coverage of remote regions and oceans where radar installations do not exist. Recent studies introduce AI techniques to detect convective initiation[24,25, 29] or make convection nowcasting[23] with geostationary satellite data. In addition, some works[30,31,32,33] such as ProbSevere-V3[31] combine satellite data and ground radar[32] or NWP results[33] for severe weather nowcasting. However, they provide relatively accurate nowcasting results only in a short period (20 minutes – 2 hours). The essential reason is the lack of stable and reliable AI architectures designed to model the spatiotemporal evolution patterns of convective clouds, especially in the reasonable simulation of stochastic and uncertain motion patterns in the temporal evolution process of convective clouds.

Inspired by the natural advantages of diffusion models in modeling stochastic processes[34], we propose a novel perspective that the stochastic motion tendency of convective clouds can be regarded as a physical diffusion process, which can be effectively modeled with forward and backward procedures of diffusion models[36]. Moreover, a recent work[35] systematically compares different diffusion models for cloud modeling and demonstrates their advantages in capturing cloud growth and decay, as well as in preserving high-resolution details. Based on the insights above, we propose a novel deep diffusion model to establish a convective nowcasting system utilizing satellite brightness temperature data and domain knowledge from meteorological experts, and conduct long-term tests and objective validation based on the FengYun-4A satellite. This forecasting system achieves, for the first time, effective convection nowcasting up to 4 hours, and features key forecasting capabilities such as broad coverage (about 20,000,000 km$^2$), high accuracy, and high spatiotemporal resolution (15 minutes; 4 km). Its performance reaches a new height in convection nowcasting compared to both the existing state-of-the-art AI models and the most advanced traditional models. Moreover, the forecasting system is highly transferable, with relevant methodologies and algorithms that can be easily transferred across different satellite platforms.

## High-Resolution Convection Nowcasting System

We develop a high-resolution convection nowcasting system with deep learning techniques. Specifically, we formulate the convection nowcasting task as two stages, namely, satellite nowcasting and convection detection. For the satellite nowcasting, we propose the DDMS to model spatiotemporal patterns of convective clouds to predict the next 4-hour satellite sequences.

For the convection detection, we adopt a combined manner of deep learning and expert knowledge to achieve the convective clouds recognition. In this article, we utilize FengYun-4A satellite data[37] from 2018 to 2021 year as training data and employ 2022 and 2023 data as the test data. The designed DDMS is trained on a GPU server with eight RTX A6000 GPUs (48GB memory) and takes two million training batches to converge (about 30 days).

As shown in **Fig. 1**, the DDMS convection nowcasting system consists of a data process module, a satellite nowcasting module, and a convection detection module. First, in the data process module, collected FengYun-4A satellite images (AGRI-L1 RGEC and DISK data, 4km, 10.8μm-band) are first temporally sampled into sequences with a 15-minute interval, and then spatially cropped into 256 × 256-size parts to reduce the GPU memory demand of the satellite nowcasting module in the training phase. Second, the designed DDMS model is trained to predict the next 4-hour satellite sequences with historical 2-hour ones as input. Here, the 2-hour history window provides sufficient temporal context for capturing the key developmental features of convective systems, which typically evolve on the scale of a few hours, while also keeping the computational cost manageable. Notably, similar settings can be found in related weather nowcasting studies[14,15] with AI techniques. Since our model is a fully convolutional network, once trained, it can be tested with arbitrary size sequences such as 1,280 × 730-size (about 20,000,000 km$^2$ coverage). Third, the convection detection module accounts for automatically recognizing the convective clouds in the predicted satellite sequences. To achieve this, we manually label a convective cloud dataset based on FengYun-4A with the guidance of experts in the National Satellite Meteorological Center and train an UNet to detect the convective clouds.

Through the three modules, we build a high-resolution convection nowcasting system. It surpasses the traditional nowcasting approaches Persistency and PySTEPS, and the AI-based nowcasting methods PredRNN-v2 and NowcastNet, in terms of multi-scaled quantitative evaluation. Besides, in the severe convective weather events, our method more accurately predicts the growth, dissipation, and intensity of convective clouds. Notably, the intensity of convective clouds here is not directly predicted. Instead, the forecasting brightness temperature is employed as a potential measure of intensity, while the convection detection module predicts the spatial coverage of convective clouds.

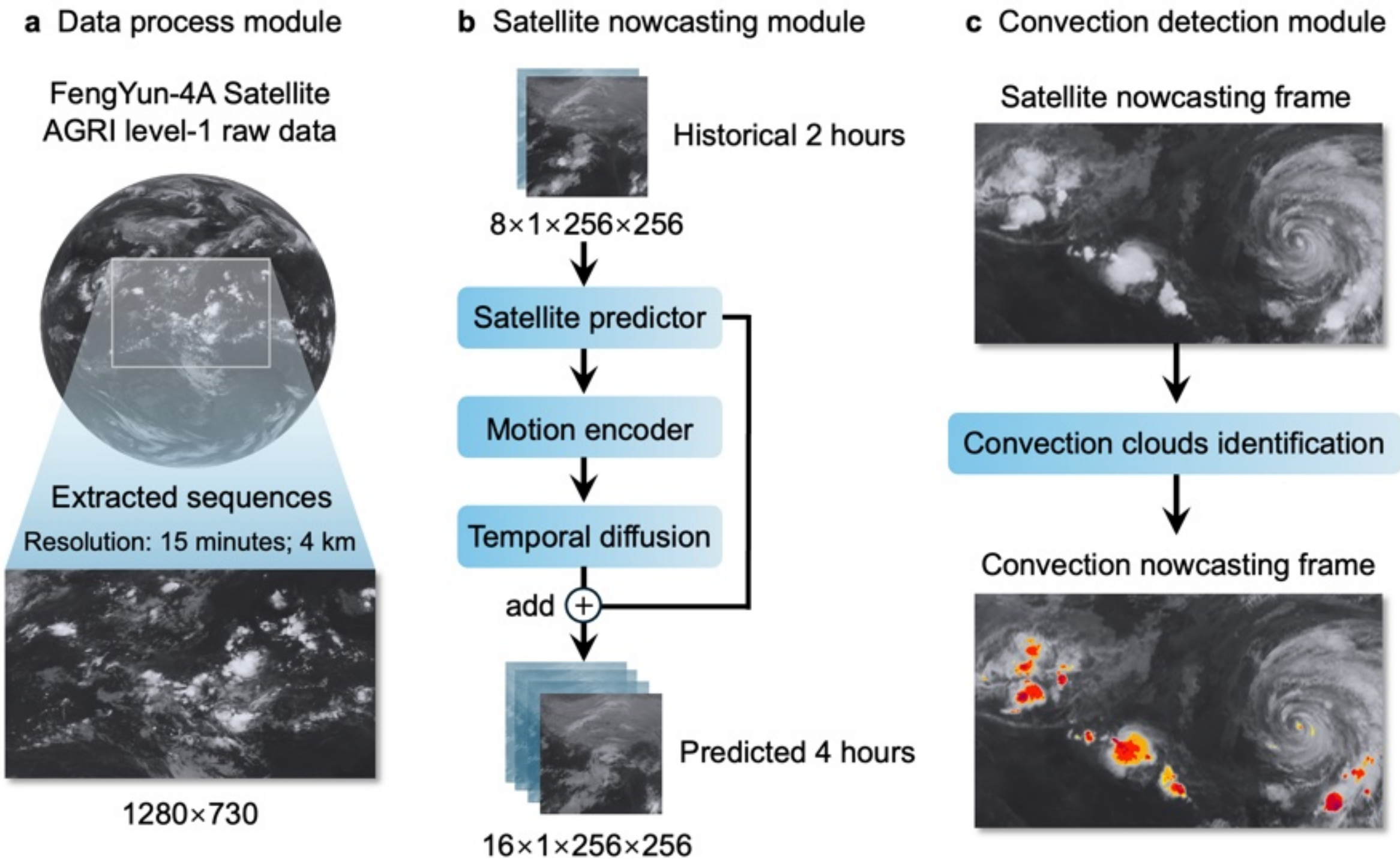

**Fig. 1 Framework of DDMS convection nowcasting system. a**, Data process module, collected FengYun-4A AGRI-L1-RGEC and DISK data are processed into regular satellite sequences through spatial cropping and temporal sampling operations. **b**, Satellite nowcasting module, predicts the next 4-hour satellite data based on the historical 2-hour satellite data. **c**, Convection detection module, identifies the convective clouds in the predicted satellite sequences.

## Results

### Evaluation Settings.

We validate the nowcasting ability of the proposed DDMS against the state-of-the-art traditional and AI-based prediction methods. We employ a classic approach Persistency[38] and an advection-based nowcasting method PySTEPS[13], as baselines of traditional methods, which have been widely used in meteorological centers worldwide[12,40]. We select PredRNN-v2[39] and NowcastNet[15] as baselines of deep learning methods. The PredRNN-v2 is a powerful spatiotemporal prediction model, and the NowcastNet delivers state-of-the-art performance for 3-hour radar nowcasting, both of which are deployed for the China Meteorological Administration for operational nowcasting. In this work, we utilize FengYun-4A satellite 10.8μm-band observations in 2018-2021 to train the above models, and employ spring and summer samples (May to August) in 2022-2023 to evaluate the nowcasting performance. During these months, the motions of convective clouds are most fierce, and the clouds' growth and decay are common in East Asia, accompanied by abrupt meteorological disasters.

The details of data processing, including the construction of the training and validation datasets during the training procedure, are described in the Data Processing Module section. For PySTEPS, we follow the standard configuration and apply the dense Lucas–Kanade optical flow method for extrapolation. For PredRNN-v2 and NowcastNet, we adopt the official open-source settings, which ensure reasonable configurations. Furthermore, all AI models are trained with an early stopping strategy to maintain a fair comparison.

## Convection Nowcasting Results.

### Quantitative Evaluation.

**Fig. 2** shows the quantitative convection nowcasting results of DDMS, NowcastNet, PredRNN-v2, PySTEPS, and Persistency, w.r.t lead times and months. The critical success index (CSI) is a key evaluation metric that measures the nowcasting accuracy of convective clouds. Here, we adopt CSI scores at different spatial scales (CSI-1, CSI-2, CSI-4, and CSI-12) to comprehensively validate the prediction results. In particular, we conduct experiments using FengYun-4A data from 2022 and 2023 to validate the prediction stability of the nowcasting methods across different years.

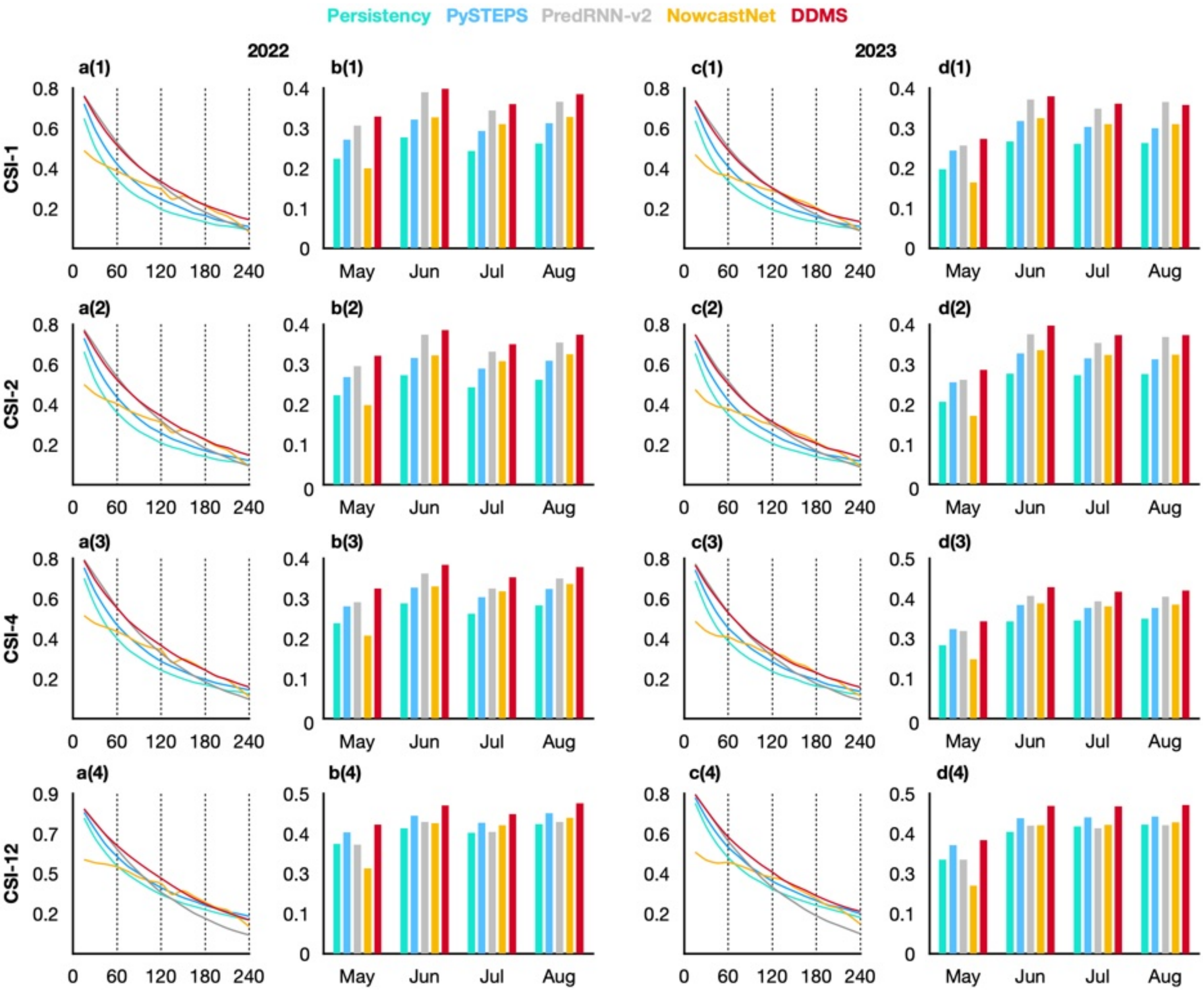

**Fig. 2 Quantitative evaluation of DDMS.** Curves of the critical success index at different spatial scales, CSI-1 (4km), CSI-2 (8km), CSI-4 (16km), and CSI-12 (48 km) with nowcasting lead time averaged from May to August in 2022 **(a(1)-a(4))** and 2023 **(c(1)-c(4))**. Histograms of the CSI scores respectively for 2022 **(b(1)-b(4))** and 2023 **(d(1)-d(4))** May, June, July, and August. Red (yellow, grey, blue, cyan) curves and bars indicate DDMS (NowcastNet, PredRNN-v2, PySTEPS, and Persistency).

First, from the perspective of forecasting lead time across the four evaluated spatial scales (4-, 8-, 16-, and 48-km), DDMS consistently outperforms traditional methods Persistency and PySTEPS (**Fig. 2 a(1)-a(4)**, **c(1)-c(4)**). This advantage is evident at all evaluated scales and is generally more pronounced at shorter lead times, although the performance gap narrows at coarser scales (16- and 48-km) during the longer forecasting hours. Second, the AI-based method NowcastNet produces competitive results over a short period (from about 120 minutes to 210 minutes), particularly at the 4-km scale, but it performs considerably worse outside this interval. Third, PredRNN-v2 shows moderate advantages during the first two hours at small scales, while its performance declines at longer lead times and at coarse scales.

Next, we analyze the forecasting abilities of DDMS and baseline methods across months and spatial scales. DDMS consistently delivers the best performance in nearly all months (**Fig. 2 b(1)-b(4)**, **d(2)-d(4)**). At the 4-km scale, the AI models NowcastNet and PredRNN-v2 generally outperform Persistency and PySTEPS, which highlights the advantages of AI techniques in modelling high-resolution convective systems. In contrast, Persistency and PySTEPS exhibit relatively improved performance at the 48-km spatial scale, indicating that advection-based approaches can effectively capture large-scale cloud patterns. Compared with the baselines, DDMS maintains a clear superiority across all scales. Finally, PredRNN-v2 shows competitive results at the 4-km scale evaluation. However, its performance declines at the 8-km, 16-km, and 48-km scale evaluations. This is mainly because PredRNN-v2 is trained under a MSE loss function, which encourages conservative and averaged predictions. Such forecasts can reduce false alarms and produce seemingly favorable quantitative results, but they inevitably weaken the representation of convective growth or dissipation at coarser scales. Specifically, **Tab. 1** presents the quantitative improvements of DDMS over PredRNN-v2 at different scales, which demonstrates the advantages of DDMS in modeling large-scale convective cloud systems.

**Tab. 1 Quantitative performance improvements of DDMS over PredRNN-v2 (%).**

| Metrics | 2022-05 | 2022-06 | 2022-07 | 2022-08 | 2023-05 | 2023-06 | 2023-07 | 2023-08 |
|---|---|---|---|---|---|---|---|---|
| CSI-1 | 7.30 | 2.10 | 4.66 | 4.82 | 6.64 | 2.35 | 3.44 | -1.75 |
| CSI-2 | 8.56 | 3.18 | 5.67 | 5.75 | 8.21 | 3.49 | 4.75 | 0.08 |
| CSI-4 | 11.64 | 6.09 | 8.33 | 8.19 | 11.70 | 6.89 | 8.19 | 4.86 |
| CSI-12 | 18.61 | 12.51 | 14.71 | 14.23 | 20.61 | 15.26 | 17.25 | 15.88 |

**Case Evaluation.**

To visually evaluate the nowcasting ability of DDMS, we report two nowcasting results of typical severe convection events in **Fig. 3** and **Fig. 4**. PySTEPS preserves sharp appearance details but fails to accurately capture the complex motion patterns of clouds, resulting in large deviations from observations. The limitations stem from the difficulty of advection-based physical equations in characterizing the non-stationary motions such as rotation, dissipation, and accumulation. Compared with PySTEPS, NowcastNet and PredRNN-v2 are able to capture the growth and dissipation of convection. In **Fig. 3**, the left cell in the white box grows during the first two hours and dissipates in the following two hours, while the right cell continuously dissipates throughout the four hours. NowcastNet significantly overestimates the growth of both cells in the first two hours and overpredicts their dissipation in the last two hours. PredRNN-v2 makes similar predictions in the first two hours, which are better than those of NowcastNet, but it still fails to forecast the dissipation of the right cell in the last two hours. In contrast, our proposed DDMS can distinguish between the growth and dissipation of different cells under varying forecasting horizons, thereby producing more accurate predictions. In **Fig. 4**, DDMS again more accurately forecasts the convection growth in the first two hours and provides better predictions of the dissipation in the last two hours. More convection nowcasting samples are provided in the supplementary file.

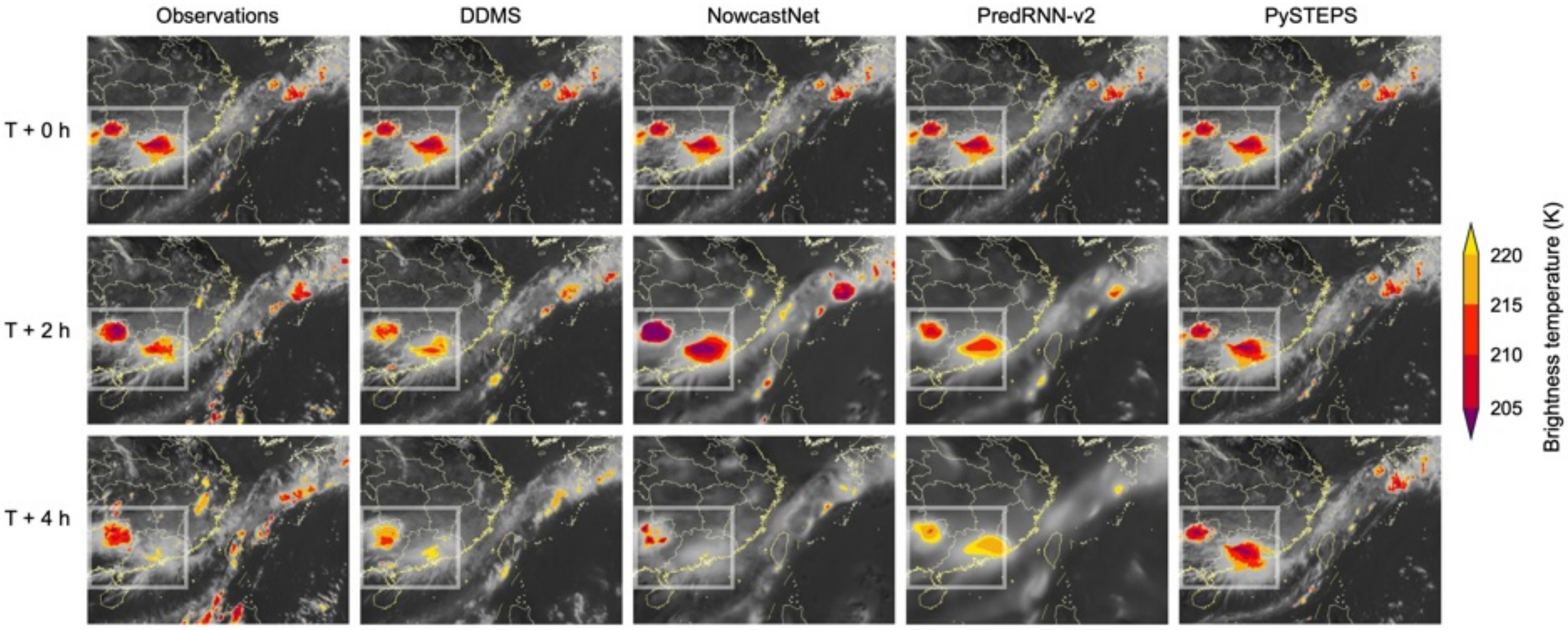

**Fig. 3 Case evaluation of a severe convection event with an extreme rainstorm starting on 16 June 2022 in South China.** This convection event is also accompanied by flood warnings of the Pearl River. The color denotes the predicted brightness temperature (K) of convective clouds. The locations of the extreme event are marked with white boxes. The samples are cropped to highlight local details. From T + 0 h to T + 2 h, NowcastNet significantly overestimates the growth of the two major cells, especially the right cell, which is actually dissipating. From T + 2 h to T + 4 h, however, it overpredicts the dissipation. PredRNN-v2 makes similar predictions in the first two hours, which are better than those of NowcastNet. However, in the last two hours, PredRNN-v2 fails to capture the dissipation of the right cell. PySTEPS is unable to predict either the growth or dissipation of the convective clouds. The three methods fail to deliver satisfactory 4-hour convection nowcasting, while our proposed DDMS produces much more accurate nowcasting results.

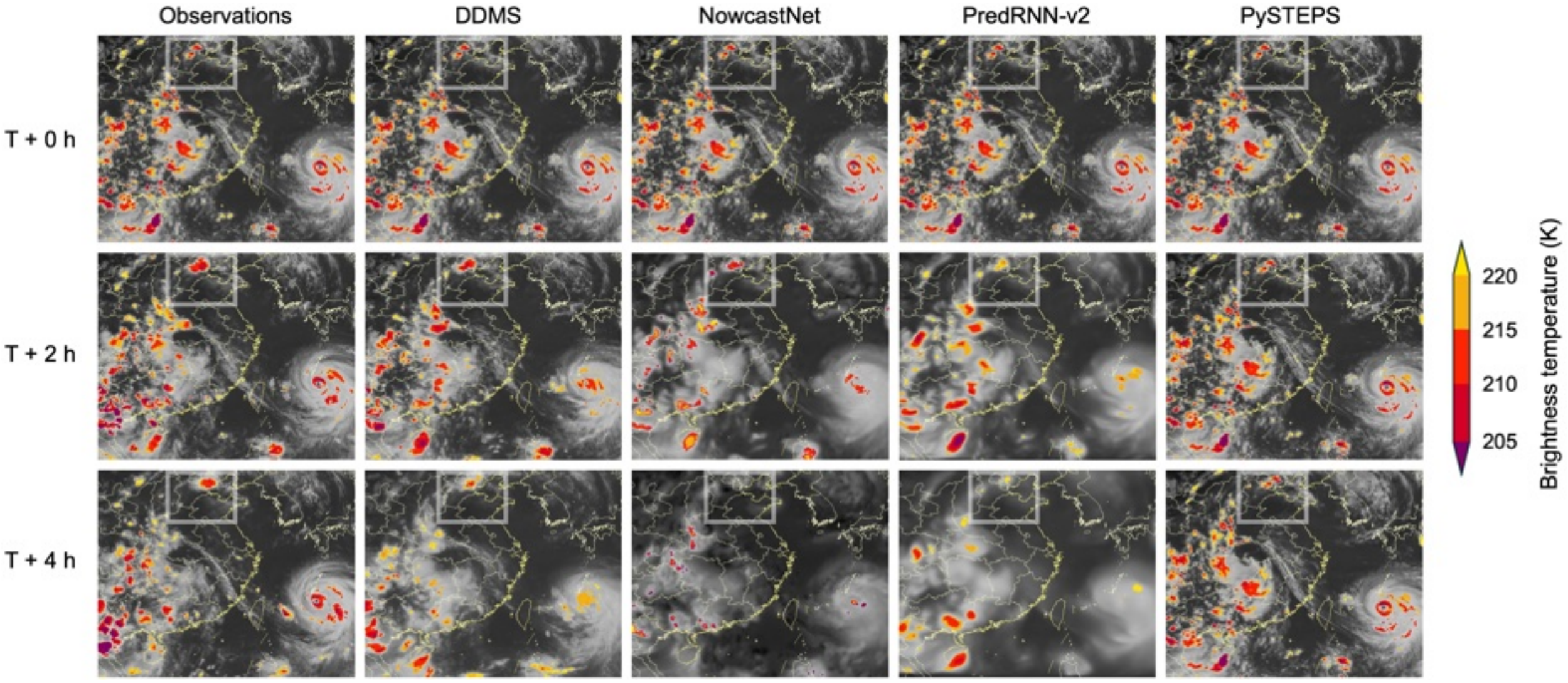

**Fig. 4 Case evaluation of a severe convection event with extreme rainstorm starting on 29 July 2023 in Beijing-Tianjin-Hebei region of China.** This convection event is affected by the typhoon Dusuirei (in the bottom-right corner). NowcastNet and PredRNN-v2 cannot provide useful prediction results at the 4-hour lead time, while PySTEPS fails to model the growth and dissipation of convective cells in the Beijing-Tianjin-Hebei region. Compared with the methods, DDMS again delivers more accurate nowcasting results.

## Discussion

In this article, we propose DDMS and build a high-resolution convection nowcasting system of satellite data. Our model is trained on 2018-2021 FengYun-4A satellite data and tested on 2022-2023 data. Compared with state-of-the-art nowcasting baseline approaches, DDMS consistently delivers the most accurate and skilful 4-hour convection nowcasting results in terms of quantitative and case evaluation.

Convection nowcasting can be further improved in some aspects. In terms of methodology, more efficient AI techniques can be explored to deliver more accurate and higher spatio-temporal resolution convection nowcasting. Besides, from the perspective of data, fusing multi-source data such as satellite and radar observations can promote the accuracy and interpretability of AI-based methods for convection nowcasting. Also, DDMS's training cost is large due to the diffusion architecture, and it is important to explore more efficient diffusion models. In addition, explicitly incorporating the conventional physical knowledge as constraints for AI-based methods is a promising direction. At last, as a diffusion-based forecasting method, DDMS can produce ensemble predictions at inference, which provides a natural way to quantify forecasting uncertainty. Though ensemble predictions are not evaluated here, we highlight this as an important potential strength of DDMS for future probabilistic convection nowcasting applications. We hope our work can inspire more researchers to investigate convection nowcasting with meteorological satellite data.

## Methods

Here we introduce the details of the data process module, satellite nowcasting module, and convection detection module in the high-resolution convection nowcasting systems, as well as the evaluation metrics and training settings.

### Data Process Module.

In this article, we collect FengYun-4A meteorological satellite AGRI-L1 data from 2018 to 2023. Data from 2018 to 2021 are utilized for training the satellite nowcasting module, while the remainder is for convection nowcasting evaluation. With a multichannel visible infrared scanning imager, FengYun-4A can produce 14-waveband satellite images, where different wavebands aim at capturing different meteorological elements such as water vapor and temperature. To achieve convection nowcasting, we select the 12$^{th}$ channel (10.8μm-band, long-wave infrared) images with 4km spatial resolution, which is suitable for observing the convective clouds in the atmosphere. The FengYun-4A satellite scans the China area (REGC) and the entire observation area (DISK) with an interval of 4-6 minutes and 15 minutes, respectively. First, we extract 24-frame sequences with an interval of 15 minutes, where each frame is 1,280 × 730-size. Second, to save the training memory demand, we crop 256 × 256-size parts of each frame and select 1% samples for constructing the validation dataset. In this manner, we obtain 441,895 sequences for training the satellite nowcasting module and 4355 sequences for validation, during the training procedure. Notably, the brightness temperature values decrease from 320 to 180, which is rescaled to [0, 255] before being represented as grayscale images. Moreover, the inputs of PredRNN-v2 are normalized to the range [0, 1], while NowcastNet doesn't adopt normalization. The inputs of DDMS are normalized to the range [-1, 1].

## Satellite Nowcasting Module.

As shown in **Extended Data Fig. 1**, the module consists of three parts, namely, a satellite predictor $P$, a motion encoder $M$, and a temporal diffusion part $D$. First, the satellite predictor accounts for producing the predicted satellite sequence $\hat{Y} = \{\hat{X}_{l+1}, \hat{X}_{l+2}, \ldots, \hat{X}_{l+k}\} \in \mathbb{R}^{k \times m \times n}$, with given the past frames $X = \{X_1, X_2, \ldots, X_l\} \in \mathbb{R}^{l \times m \times n}$. $m$ and $n$ denote the width and length of each satellite image, respectively. Second, the temporal diffusion part is specially designed to model the temporal evolution dynamics of divergence between the predicted sequence $\hat{Y}$ and observation sequence $Y = \{X_{l+1}, X_{l+2}, \ldots, X_{l+k}\}$. Third, the motion encoder is designed to capture the long-term motion tendency of the predicted sequence $\hat{Y}$, to guide the denoising process of the temporal diffusion part. By combining the predicted sequence by satellite predictor $P$ and the reconstructed difference sequence by temporal diffusion $D$, we obtain the final prediction sequence $\hat{Y}_{final} = P(x) + D(Y - \hat{Y})$. Next, we introduce details of the three parts and the training loss of the satellite nowcasting module.

**Satellite Predictor.** Overall, the satellite predictor follows a classic RNN architecture ConvGRU[41]. It is a simple yet effective spatiotemporal prediction baseline, whose key formulae are defined as follows:

$$
\begin{aligned}
z_t &= \sigma(W_{xz} * \mathcal{X}_t + W_{hz} * \mathcal{H}_{t-1} + b_z), \\
r_t &= \sigma(W_{xr} * \mathcal{X}_t + W_{hr} * \mathcal{H}_{t-1} + b_r), \\
\widetilde{\mathcal{H}}_t &= \tanh(W_{xh} * \mathcal{X}_t + W_{hh} * (r_t \circ \mathcal{H}_{t-1}) + b_h, \\
\mathcal{H}_t &= (1 - z_t) \circ \mathcal{H}_{t-1} + z_t \circ \widetilde{\mathcal{H}}_t.
\end{aligned}
\tag{1}
$$

The $z_t$ and $r_t$ denote the update gate and reset gate, respectively. The $\widetilde{\mathcal{H}}_t$ and $\mathcal{H}_t$ represent the candidate hidden state and final hidden state, respectively. The $*$ is the convolution operation, the matrices like $W_{xz}$ and $W_{hz}$ are corresponding weights, and terms like $b_z$, $b_r$, and $b_h$ are biases. $\sigma$ and tanh are activation functions, and the symbol $\circ$ denotes the Hadamard multiplication.

Based on the classic ConvGRU backbone, the satellite predictor introduces the deep residual mechanism and skip connection manner to enhance spatiotemporal modeling ability. Specifically, it contains an encoder part and a decoder part. In the encoder, three ResNet-GRU-Conv block layers and one ResNet-GRU block layer are utilized to encode multi-level spatiotemporal dynamics of satellite sequences. In the decoder, three ResNet-DeConv block layers and one convolutional layer with a kernel size of 1 are employed to produce predicted frames by decoding the multi-level spatiotemporal features. Notably, in each time step, different level hidden states generated by the GRU units in the encoder are passed to the decoder in a skip connection manner, which can better preserve spatial details. In an autoregressive manner, the satellite predictor produces the k-length predicted sequence.

**Motion Encoder.** The motion encoder adopts the previous four layers of a satellite predictor to encode the predicted sequence to obtain different level motion patterns at each time step, which are utilized to guide the denoising process of the temporal diffusion part. In particular, the motion encoder does not share parameters with the satellite predictor, and is removed the skip connection operations. The encoding procedure is formally defined as follows:

$$\text{Motions}^{i} = M(\hat{Y}_i), \tag{2}$$

$$\text{Motions}^{i} = \{\ \text{Motion}_1^{i},\ \text{Motion}_2^{i},\ \text{Motion}_3^{i},\ \text{Motion}_4^{i}\}. \tag{3}$$

The $\text{Motion}_1^{i}$, $\text{Motion}_2^{i}$, $\text{Motion}_3^{i}$, and $\text{Motion}_4^{i}$ denote different level motion patterns at the *i*-th time step.

**Temporal Diffusion.** With the guidance of the motion encoder, the temporal diffusion part is proposed to model the temporal evolution of the difference satellite sequence. To train the diffusion network, we adopt a training paradigm provided by DDPM[36], which treats the diffusion process as a first-order Markov-chain procedure. Specifically, in the forward process, Gaussian noise is gradually added to destroy the original data $x_0$, then we obtain $T$ different states $\{x_1, x_2, \ldots, x_T\}$, which is formally defined as follows:

$$q(x_{1:T} \mid x_0) = \prod_{t=1}^{T} q(x_t \mid x_{t-1}), \tag{4}$$

$$q(x_t \mid x_{t-1}) = \mathcal{N}\left(x_t \mid \sqrt{\alpha_t} x_{t-1}, (1-\alpha_t)\mathbf{I}\right). \tag{5}$$

$1-\alpha_t$ is the predefined covariance hyperparameter at the t-th time step. $\alpha_t \in (0,1), t \in \{1,2,\ldots,T\}$, and $\alpha_1 > \alpha_2 > \cdots > \alpha_T$. Through formulae (4) and (5), $t$-th state $x_t$ can be obtained in one step as folows:

$$q(x_t \mid x_0) = \int q(x_{1:t} \mid x_0) dx_{1:(t-1)} = \mathcal{N}\left(x_t \mid \sqrt{\bar{\alpha}_t} x_0, (1-\bar{\alpha}_t)\mathbf{I}\right), \tag{6}$$

where $\bar{\alpha}_t = \alpha_1 \alpha_2 \ldots \alpha_t$. When $t \to \infty$, then $\bar{\alpha}_t \to 0$, and we can get that $x_t \to \mathcal{N}(0, \mathbf{I})$.

In the reverse process, we aim to recover the original data $x_0$ from the Gaussian noise $x_T$. Formally, the denoising process is defined as follows:

$$p(x_0) = \int p(x_{0:T}) dx_{1:T}, \tag{7}$$

$$p(x_{0:T}) = p(x_T) \prod_{t=1}^{T} p(x_{t-1} \mid x_t), \tag{8}$$

$$p(x_{t-1} \mid x_t) = \mathcal{N}(x_{t-1} \mid \mu(x_t, t), \sigma_t \mathbf{I}). \tag{9}$$

The $p(x_T) = \mathcal{N}(0, \mathbf{I})$ is known, and the $\sigma_t$ is fixed and computed by $\bar{\alpha}_t$. The mean value $\mu(x_t, t)$ needs to be learned and depends on Gaussian noise. For this, DDPM adopts a UNet model $f_\theta$ to predict the noise $\epsilon$. In this work, we employ the multi-level motion patterns at the $i$-th time step $\text{Motions}^i$, as a condition to help the UNet to predict the noise. Hence, the training loss function of the denoising process is as follows:

$$L(\theta) = \mathbb{E}_{\mathbf{x}_0, t, \epsilon} \left\| \epsilon - f_\theta\left(x_t, t, \text{ Motions}^i\right) \right\|^2, \tag{10}$$

$$x_t = \sqrt{\bar{\alpha}_t} x_0 + \sqrt{1 - \bar{\alpha}_t} \epsilon. \tag{11}$$

**Training Loss.** We employ a predicted loss, namely, mean absolute error loss, to train the satellite predictor. We use a diffusion loss to train the temporal diffusion part. By combining the two loss terms, we obtain the training loss of the satellite nowcasting module as follows:

$$L(\hat{Y}, Y) = \sum_{i=1}^{k} \left( \left| Y_i - \hat{Y}_i \right| + \lambda * Diff\left(Y_i - \hat{Y}_i\right) \right). \tag{12}$$

The coefficient $\lambda$ is utilized to balance the two loss terms. $Diff(.)$ is implemented by optimizing formulae (10) and (11), where the initial state $x_0 = Y_i - \hat{Y}_i$. The forward training process and the backward testing process of DDMS are summarized as Algorithm 1 and Algorithm 2, respectively.

Algo. 1 The training procedure of DDMS

**Input:** $l$-frame historical sequence $X = \{x_1, x_2, \dots, x_l\}$; $k$-frame future sequence $Y = \{x_{l+1}, x_{l+2}, \dots, x_{l+k}\}$

**Output:** The optimal parameters $\theta$ of DDMS

**Setting:** Initialize parameters $\theta$ of DDMS; initialize the predefined covariance hyperparameter $\alpha$; set sample step $T$; set learning rate $l$ and weight coefficient $\lambda$; Assume $P$, $M$, and $D$ represent Satellite Predictor, Motion Encoder, and Temporal Diffusion, respectively.

**while** not coverage **do**

random add-noise step $t \sim \mathcal{U}(0,1,2,\cdots,T)$

The initial training loss $L = 0$

**foreach** $i = 1,2,\dots,k$ **do**

$\epsilon \sim \mathcal{N}(0, \mathbf{I}), \hat{Y}_i = P(\hat{Y}_{i-1}, X), x_0 = Y_i - \hat{Y}_i$

$x_t = \sqrt{\bar{\alpha}_t} x_0 + \sqrt{1 - \bar{\alpha}_t \epsilon}$

$L = L + \left|Y_i - \hat{Y}_i\right| + \lambda \times \| \epsilon - f_\theta(\mathbf{x}_t, t, \text{Motions}^i) \|^2$

**end**

Update parameters $\theta$ of DDMS:

$\theta = \theta - l \times \frac{\partial L}{\partial \theta}$

**end**

Algo. 2 The test procedure of DDMS

**Input:** $l$-frame historical sequence $X = \{x_1, x_2, \dots, x_l\}$; the denoise step $d$

**Output:** The $k$-frame predicted sequence $F = \{F_1, F_2, \dots, F_k\}$

**foreach** $i = 1,2,\dots,k$ **do**

$\hat{Y}_i = P(\hat{Y}_{i-1}, X)$

$x_N \sim \mathcal{N}(0, \mathbf{I})$

**foreach** $(j, j-1) \in \left\{\left(T, T - \frac{T}{d}\right), \left(T - \frac{T}{d}, T - 2 \times \frac{T}{d}\right), \dots, \left(\frac{T}{d} + 1, 1\right)\right\}$ **do**

$\epsilon_j = f_\theta(x_j, j, \text{ Motions}^i)$

$x_{j-1} = \sqrt{\bar{\alpha}_{j-1}} \left( \frac{x_j - \sqrt{1 - \bar{\alpha}_j} \epsilon_j}{\sqrt{\bar{\alpha}_j}} \right) + \sqrt{1 - \bar{\alpha}_{j-1}} \epsilon_j$

**end**

$F_i = \hat{Y}_i + x_1$

**end**

## Convection Detection Module.

Traditional convection detection on satellite images usually relies on a specific threshold value to judge convective clouds, in which the threshold value is obtained according to meteorologists' experiential judgment. The threshold detection approach is simple but not very robust, and its accuracy is also a main concern. To automatically detect the convective clouds in the predicted satellite sequence, our team and experts in the National Satellite Meteorological Center manually label a convection dataset. Specifically, we label the convective clouds by combining four principles, namely, the basic characteristics of convective clouds summarized by experts, satellite images in visible channels, the brightness temperature of satellite data, and corresponding radar maps. The detailed label principles are described in the appendix. Then we utilize the convection dataset to train a detection model (2D-UNet[42]). Finally, the convection nowcasting results are obtained by segmenting each frame of the predicted satellite sequence produced by the satellite nowcasting module with the pretrained detection model.

## Evaluation Metrics.

For quantitative evaluation, we utilize the widely adopted metrics in weather nowcasting, the critical success index at different spatial scales: CSI-1, CSI-2, and CSI-4, to evaluate the convection nowcasting results of 4-km, 8-km, and 16-km resolutions, respectively. The CSI scores denote the overall evaluation of the hit rate and false alarm for convection nowcasting. The CSI-1 is calculated from the pixel-wise images, while the CSI-2 and CSI-4 are computed from the average downsampled images with a ratio of 2 and 4, respectively. The CSI score is formally defined as follows:

$$CSI = \frac{TP}{TP + FN + FP}. \tag{13}$$

Through the convection detection module, the predicted satellite sequences $\hat{Y}_{final}$ are binarized to the convection sequences $\hat{C}$. Then we can obtain the true positives (TP, prediction = 1, target = 1), false positives (FP, prediction = 1, target = 0), and false negatives (FN, prediction = 0, target

= 1), to calculate the CSI-1 score as the formula (13). By downsampling the convection results, we can obtain the CSI-2 and CSI-4 scores in a similar manner.

### Implementation Details.

Our model DDMS is trained on a GPU server with a CPU of Intel(R) Xeon(R) Gold 6130 CPU @ 2.10GHz, 256G memory, and 8 RTX A6000 cards (48G memory). We utilize the Adam optimizer[43] to train the model, and use the exponential moving average (EMA) strategy[44] with a decay weight of 0.99 to update the parameters. The batch size and learning rate are set to 8 and 5e-5, respectively. The weight $\lambda$ in the combined training loss is 10. The dimensions of hidden states in the four-layer GRU are set to 64, 128, 192, and 256, respectively. As for the diffusion part, the forwarding step is set to 1000, while the denoising step is set to 200 with a DDIM[45] sampling manner. The details and analysis of our model are shown in the supplementary material.

## Data Availability

The FengYun-4A AGRI-4km-L1 satellite data (about 18T) can be free downloaded from the website [http://data.nsmc.org.cn/portalsite/default.aspx]. The satellite data comprises two types: DISK and REGC. DISK data covers the entire disk region, while REGC data specifically focuses on the China area. We combine these two types of data and extract sequences of 24 frames at 15-minute intervals. In addition, the test satellite data and corresponding labeled convection data will be provided soon.

## Code Availability

In this work, we utilize pytorch [https://pytorch.org/], a widely used deep learning framework, to train our model. We use Python [https://www.python.org/] to complete the data process and results evaluation. In particular, we use the h5py to load the FengYun-4A HDF files and employ opencv-python to read and save the processed satellite images. The quantitative metric CSI is calculated on the Python library Numpy. Besides, we utilize the PySTEPS [https://pysteps.github.io/], PredRNN-v2[https://github.com/thuml/predrnn-pytorch], and NowcastNet [https://codeocean.com/capsule/3935105/tree/v1] as baselines to achieve convection nowcasting. The prediction samples of DDMS over different months are provided in [https://github.com/bigfeetsmalltone/DDMS]. The source codes of DDMS and its pretrained weight will be provided soon. The source codes of the convection detection model are available from the corresponding author upon request.

## Acknowledgements

This work was supported by NSFC under Grants 62376072, 62272130, and Shenzhen Science and Technology Program under Grant JCYJ20200109113014456 and Grant KCXFZ20240903093006009, Guangdong Basic and Applied Basic Research Foundation (2023A1515110811, J.F.), and FengYun Application Pioneering Project under Grant FY-APP-ZX-2022.0220.

## Author contributions

Kuai Dai, Xutao Li, Yunming Ye, and Jingsong Wang conceived and designed the work. Kuai Dai, Xutao Li, Yunming Ye, Demin Yu, and Hui Su discussed details of the designed method and experimental results. Kuai Dai achieved the coding for the data process, model training, and results evaluation and completed manuscript writing. Junying Fang and Hui Su conducted investigation, visualization, and revision of the manuscript. Di Xian, Danyu Qin, and Jingsong Wang constructed the convection data and participated in data processing and case analysis. All the authors contributed edits.

## Competing interests

The authors declare no competing interests.

## Additional information

### Supplementary information

Supplementary Information is available for this paper.

### Correspondence and requests for materials

Correspondence and requests for materials should be addressed to lixutao@hit.edu.cn.

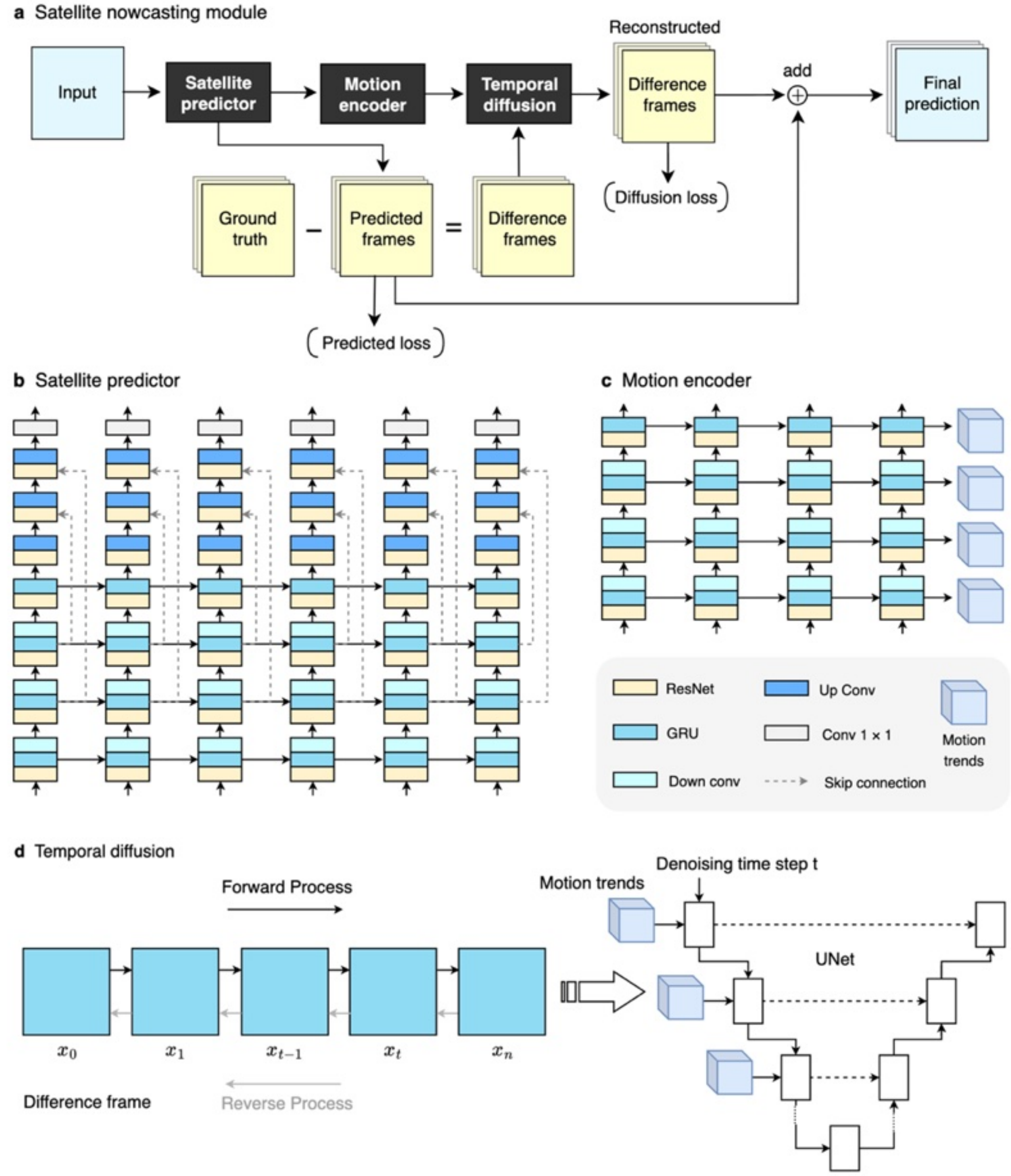

**Extended Data Fig. 1: Pipelines of the satellite nowcasting module. a**, The overall layout of the satellite nowcasting module, which contains a satellite predictor, a motion encoder, and a temporal diffusion part. The satellite predictor accounts for producing the predicted sequence with given historical frames. With the guidance of the motion encoder, the temporal diffusion part aims to model temporal evolution dynamics of difference sequences between the prediction and observation. The final prediction is obtained by combining the predicted sequence and the reconstructed difference sequence. **b**, The architecture details of the satellite predictor, which contains multiscale GRU units. **c**, The architecture details of the motion encoder, which is an encoder part of the satellite predictor and has independent parameters. **d**, The workflow of the temporal diffusion part for *i*-th difference satellite frame. The conditional UNet is employed to predict a noise in the reverse process.

# Supplementary Information: Four-hour thunderstorm nowcasting using a deep diffusion model for satellite data

Kuai Dai, Xutao Li*,Junying Fang, Yunming Ye*, Demin Yu, Hui Su, Di Xian, Danyu Qin, Jingsong Wang*

In the appendix, we provide a comprehensive analysis of our proposed method DDMS, and present more method details.

## 1. 2-hour quantitative nowcasting results.

To intuitively analyze 2-hour nowcasting abilities, we present the 2-hour prediction results in Figure 1. PredRNN-v2 and the proposed DDMS significantly outperform the other baselines. Moreover, PredRNN-v2 shows very competitive performance to DDMS, even slightly surpassing DDMS at the 4km and 8km scale evaluations. This observation is reasonable since the PredRNN-v2 models spatiotemporal patterns of convective clouds under the supervision of mean square error (MSE) or mean absolute error (MAE) loss functions. Under this scheme, PredRNN-v2 tends to generate averaged results and avoid false alarms, which improves the CSI score. However, during the latter half of the 4-hour evaluations, the performance of PredRNN-v2 declines significantly (see the a and c columns in Figure 2 of the main paper). This occurs because the uncertainty increases considerably, and such a conservative prediction approach leads to unreliable results.

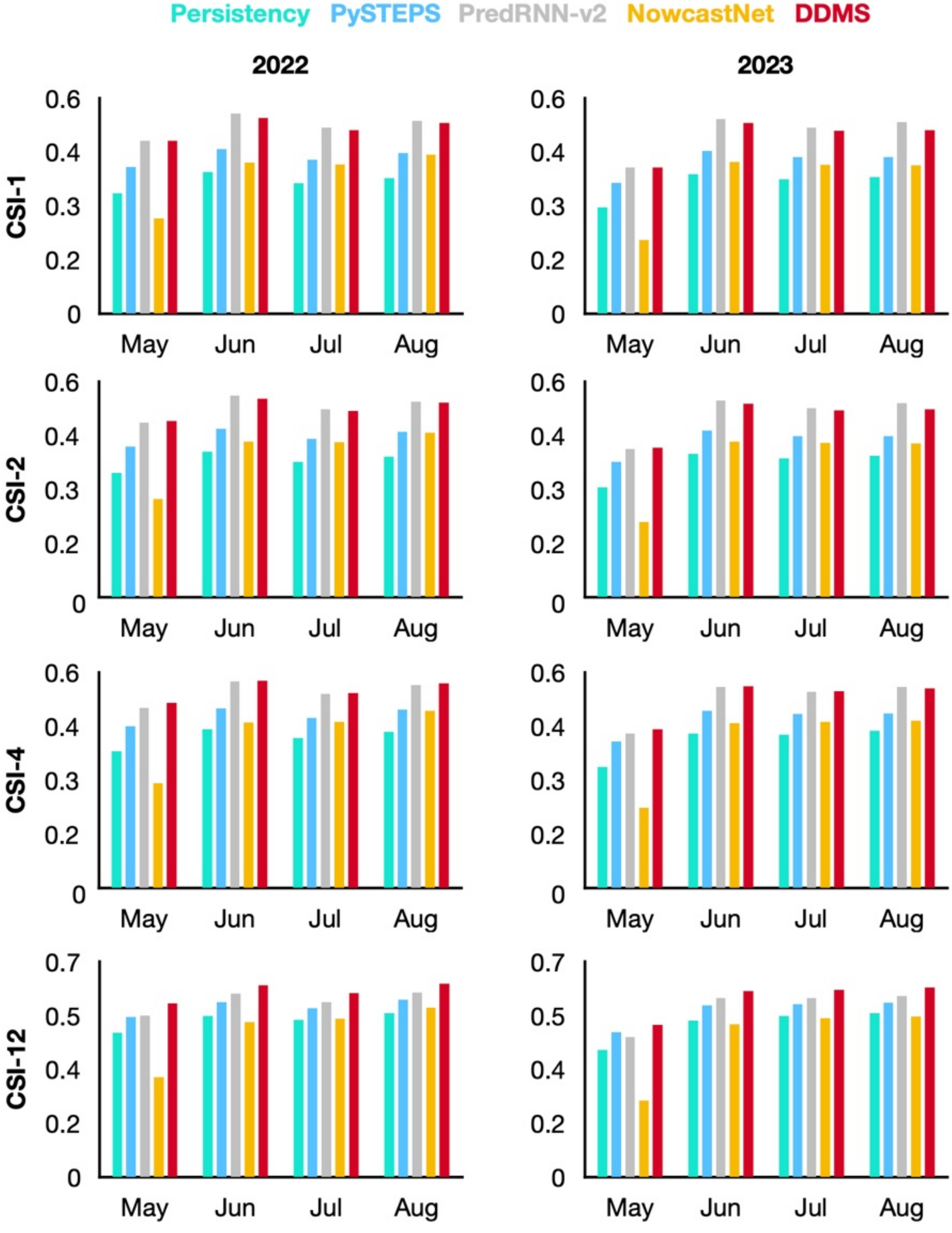

Figure 1: The 2-hour convection nowcasting results of DDMS and baselines across different months. CSI-1, CSI-2, CSI-4, and CSI-12 denote the evaluation scores at 4km, 8km, 16km, and 48km spatial scales, respectively.

## 1. More Convection Samples

Here, we report more convection nowcasting samples in Figures 2 to 9, where the motion patterns of convective clouds are very fierce. In particular, we utilize red boxes to highlight the rapid growth areas in Figures 8 and 9. By analyzing the prediction results of DDMS and the baseline models, we have the following findings here. First, PySTEPS makes sharp predictions in appearance details for all samples. However, it cannot effectively predict the formation and dissipation movements of convective clouds and delivers inaccurate predictions. Specifically, PySTEPS overestimates the region and intensity of convective clouds in Figures 2, 3, 5, and 6, while it underestimates the development tendency of convective clouds in Figures 4, 7, 8, and 9. Second, NowcastNet makes more accurate predictions in the locations of convective clouds compared to PySTEPS. However, this deep learning model largely underestimates the intensity of predicted convective clouds, and cannot make useful predictions at the 4th hour on all samples. Third, the classic AI approach PredRNN-v2, shows better performance in predicting the intensity of convective clouds, while it produces very blurry results and also suffers from the intensity degrading issue, especially at the 4-hour time step. At last, we specifically analyze the forecasting ability for rapid growth. In Figure 8, there are two main cells that grow rapidly. Our model DDMS accurately predicts the growth of the upper cell, while its prediction for the rapid growth of the bottom cell is less accurate, particularly from t + 3 h to t + 4 h. In Figure 9, DDMS captures the overall growth, although it cannot make very accurate predictions from t + 3 h to t + 4 h. Compared with the dissipation, accurately forecasting rapid growth is more challenging. Though DDMS cannot completely address this, it still outperforms both AI-based and traditional methods. Overall, compared to the three baseline approaches, our proposed DDMS models the formation and dissipation patterns more accurately and better predict the intensity and locations of convective clouds during the 4 hours. All the observations demonstrate the effectiveness of DDMS for modeling active convective clouds.

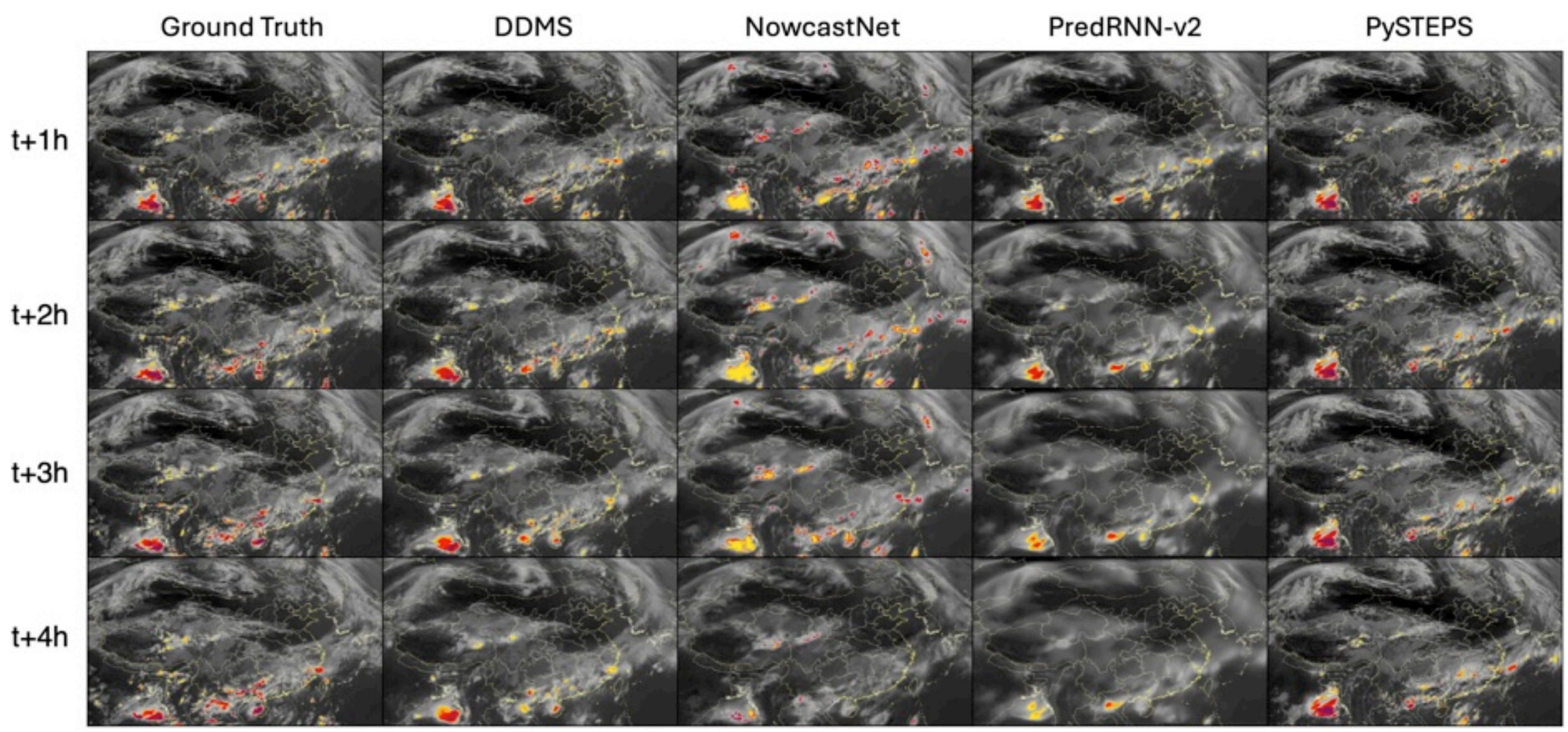

Figure 2: The 4-hour convection nowcasting sample 1, which is started at May 30, 2022 06:00 (UTC).

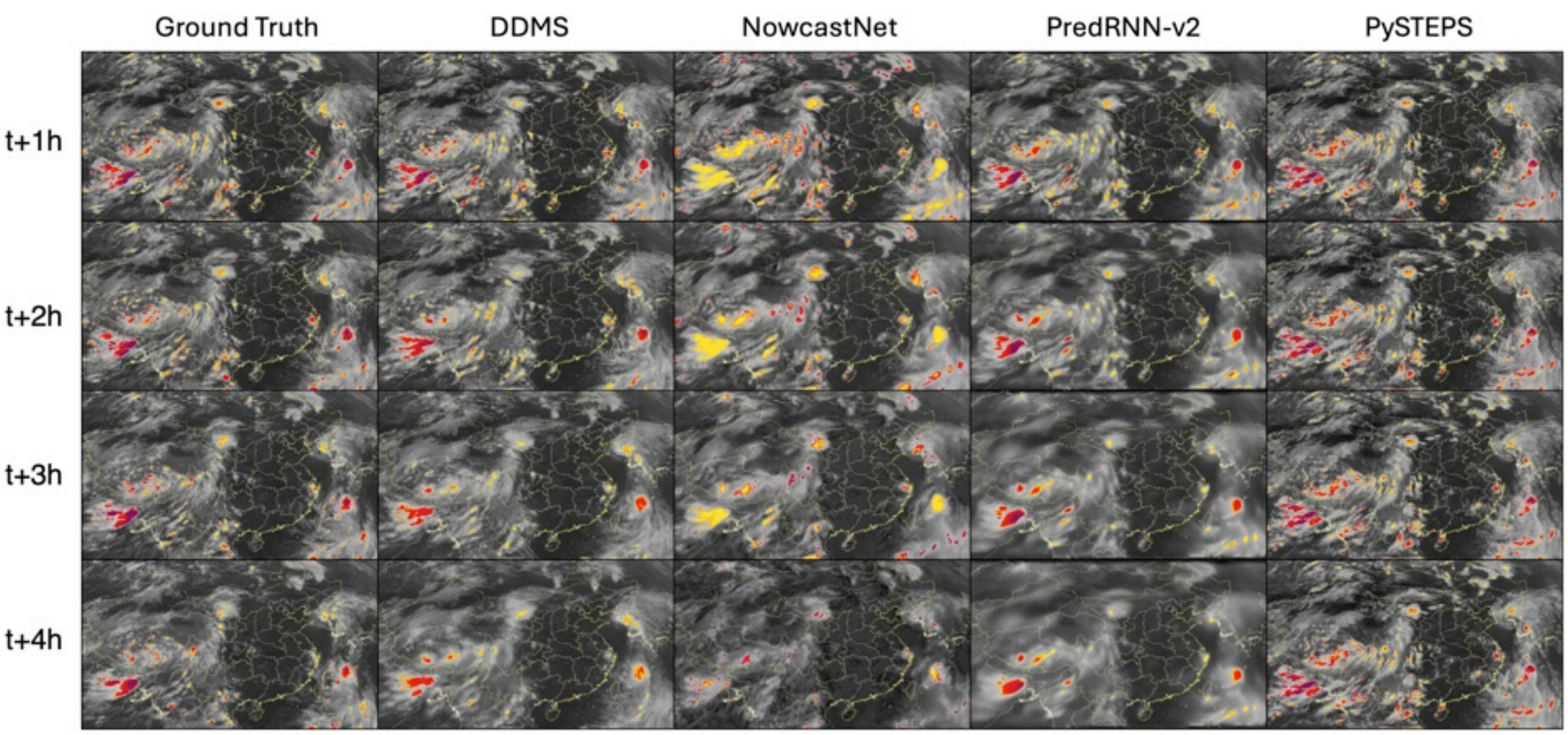

Figure 3: The 4-hour convection nowcasting sample 2, which is started at July 30, 2022 10:45 (UTC).

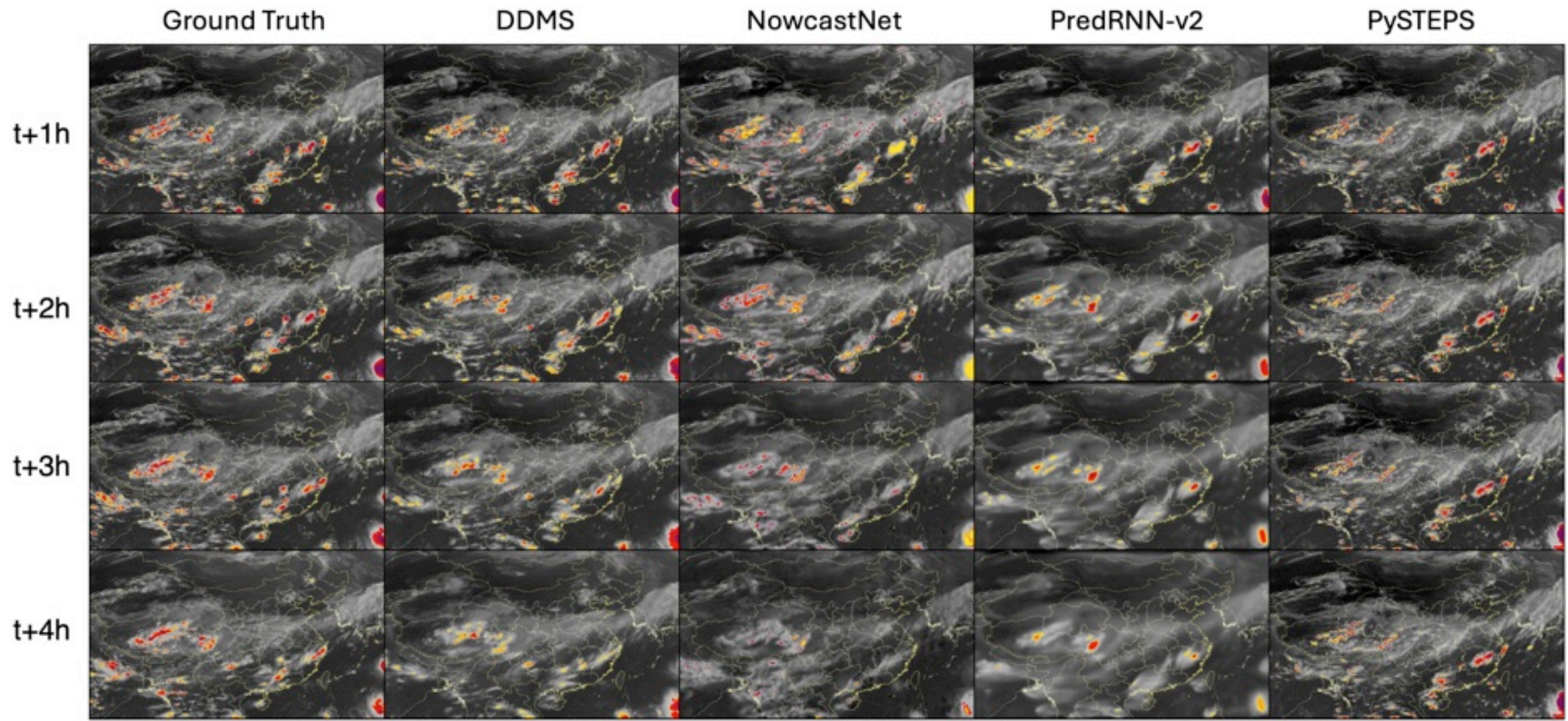

Figure 4: The 4-hour convection nowcasting sample 3, which is started at August 30, 2022 07:45 (UTC).

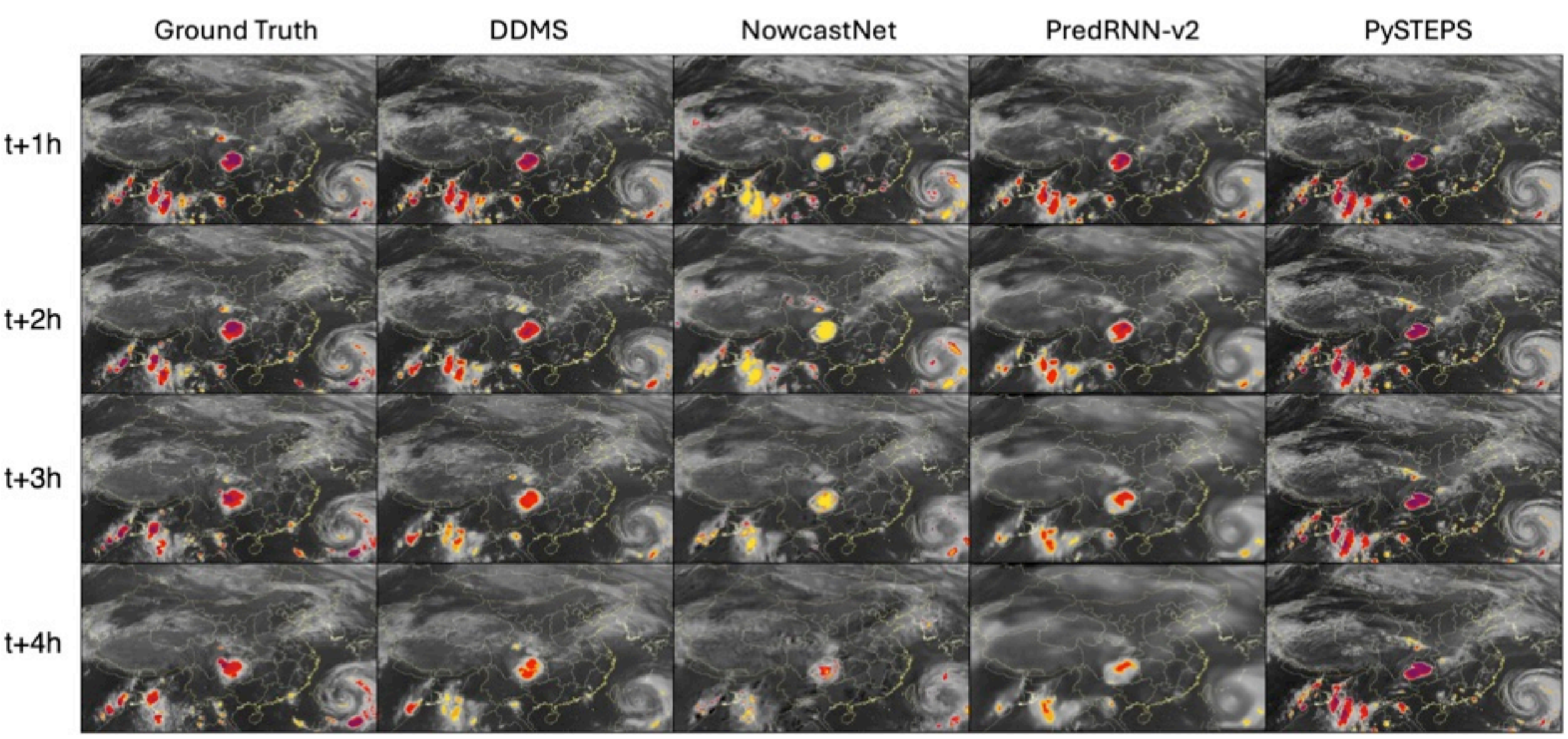

Figure 5: The 4-hour convection nowcasting sample 4, which is started at May 31, 2023 11:15 (UTC).

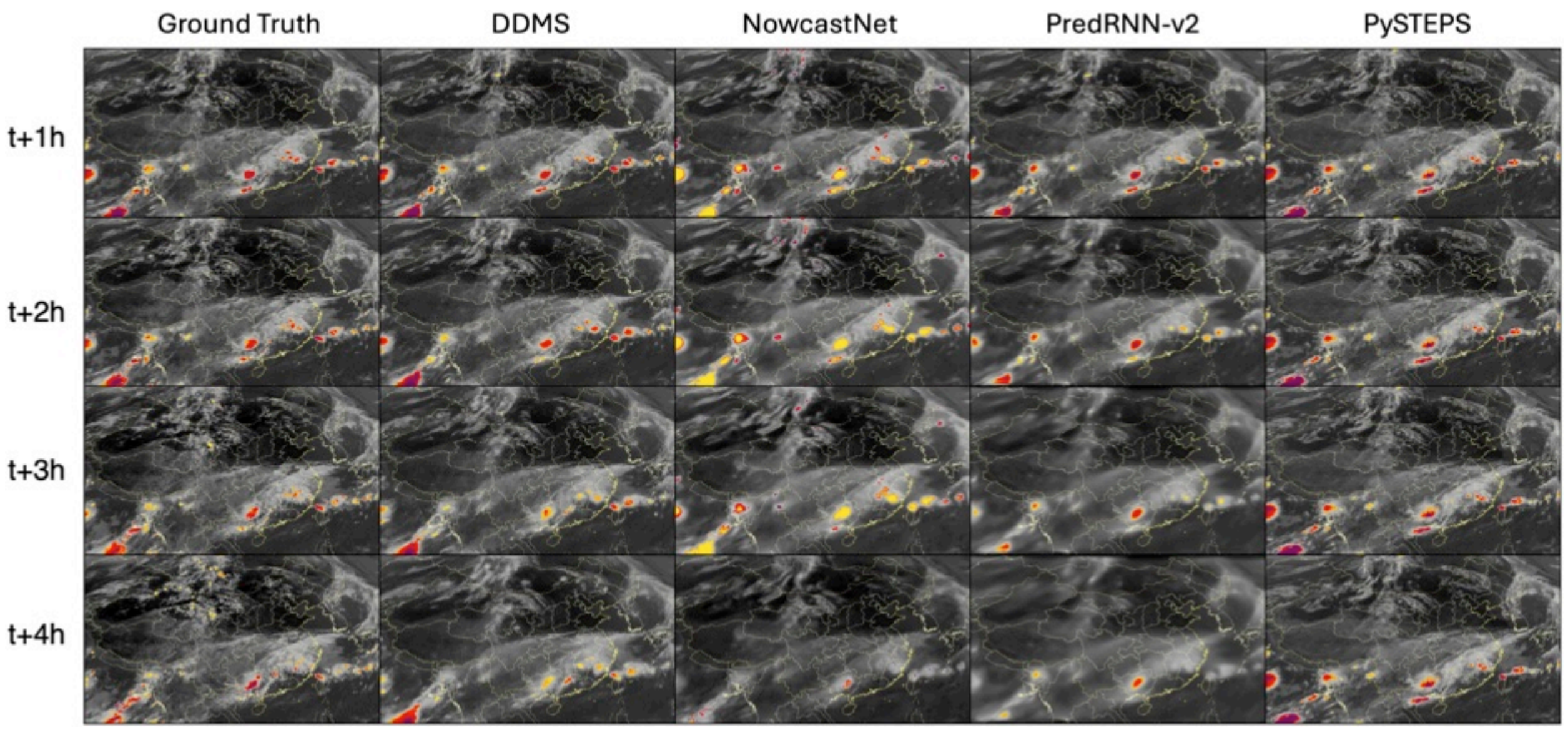

Figure 6: The 4-hour convection nowcasting sample 5, which is started at June 22, 2023 02:00 (UTC).

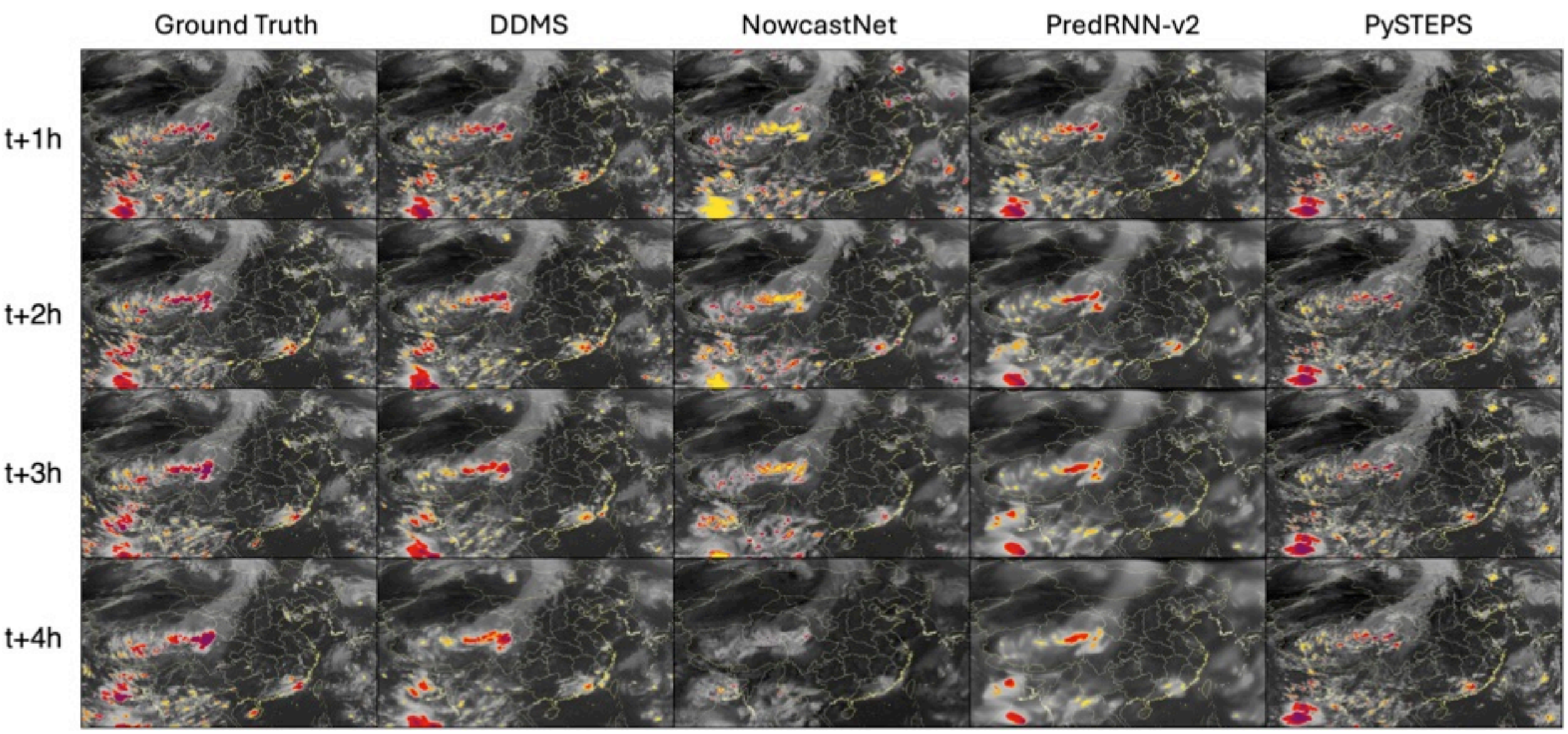

Figure 7: The 4-hour convection nowcasting sample 6, which is started at August 16, 2023 07:45 (UTC).

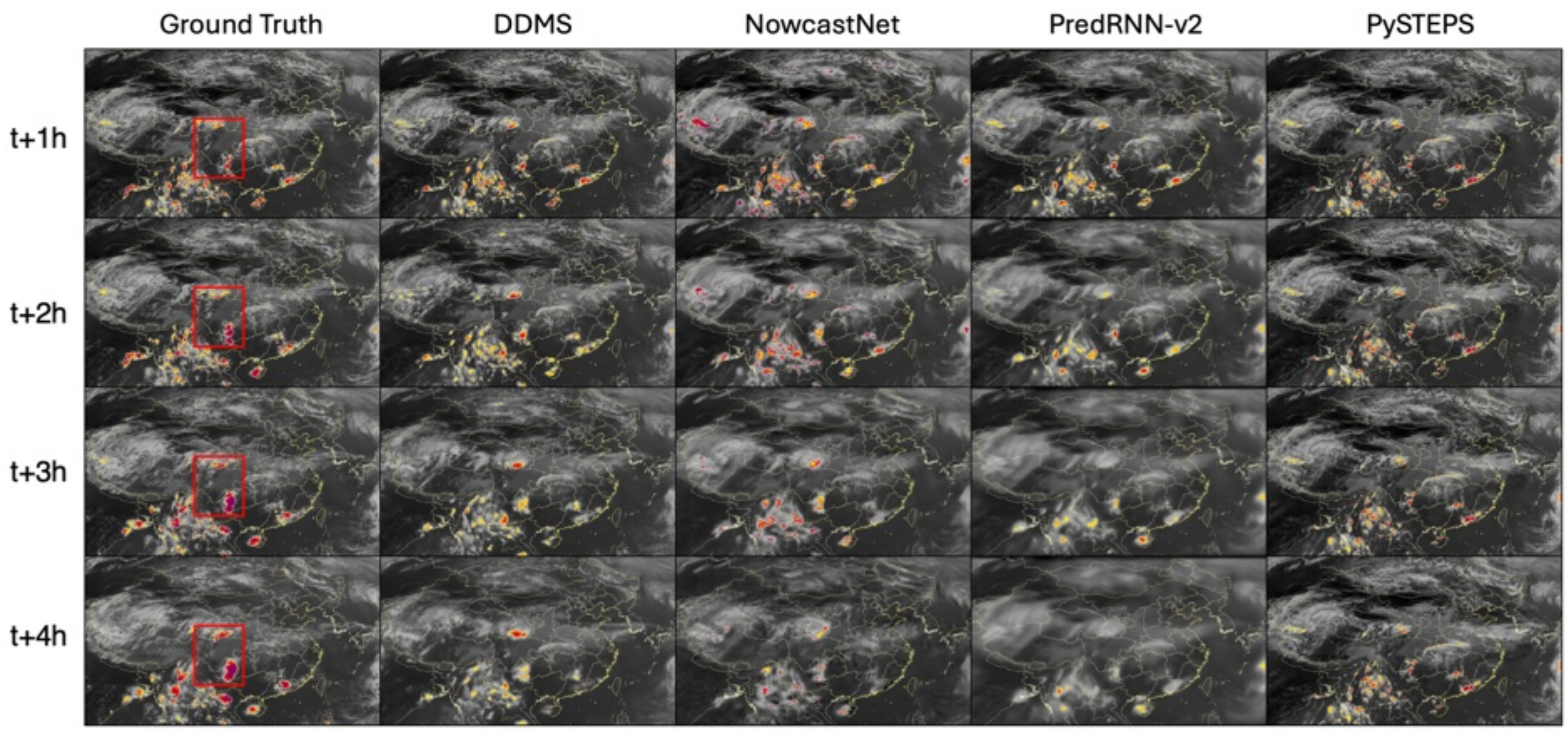

Figure 8: The 4-hour convection nowcasting sample 7, which is started at June 22, 2023 02:00 (UTC). The red box is employed to highlight the rapid growth area.

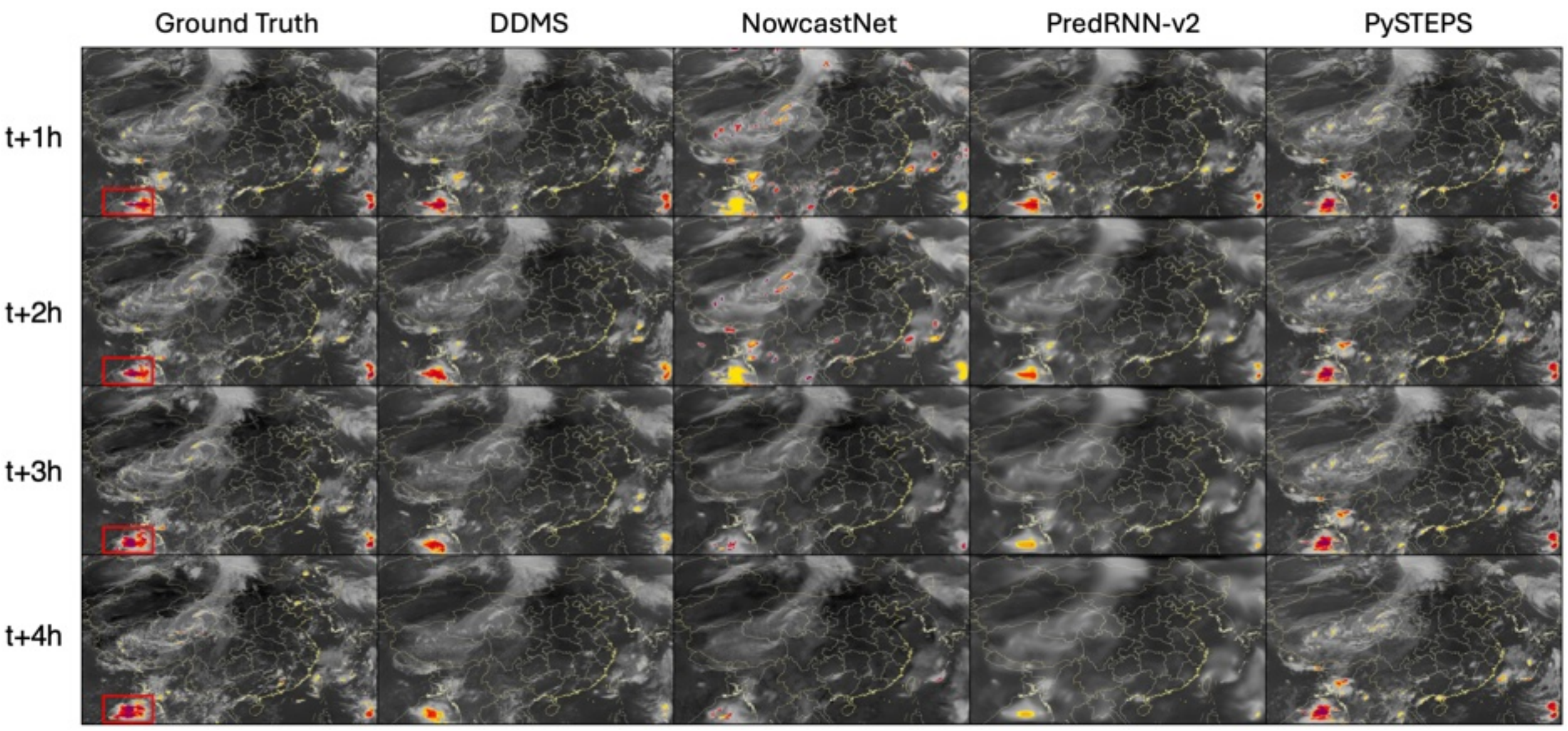

Figure 9: The 4-hour convection nowcasting sample 8, which is started at August 16, 2023 01:45 (UTC). The red box is employed to highlight the rapid growth area.

## 2. Analysis on DDMS

### 2.1 Training Stability of DDMS

To demonstrate the training stability of the proposed DDMS, we present the loss curves w.r.t the training steps in Figure 9. Intuitively, all the loss curves are steadily declining during the 1M training steps. The observation indicates the training stability of DDMS and its key components. Notably, in Figure 10, we observe that the MAE loss of the satellite predictor is increasing in the early stage. This is reasonable since the prediction ability of the satellite predictor is weak in the early stage. This causes the difference sequence to be hardly reconstructed by the temporal diffusion part with limited conditional information. Hence, the satellite predictor and temporal diffusion part engage in a co-adversarial training method at the early stage. However, as the training step increases, the satellite predictor can provide more useful guidance for the temporal diffusion part. As a result, both the losses of the two modules begin to steadily decline.

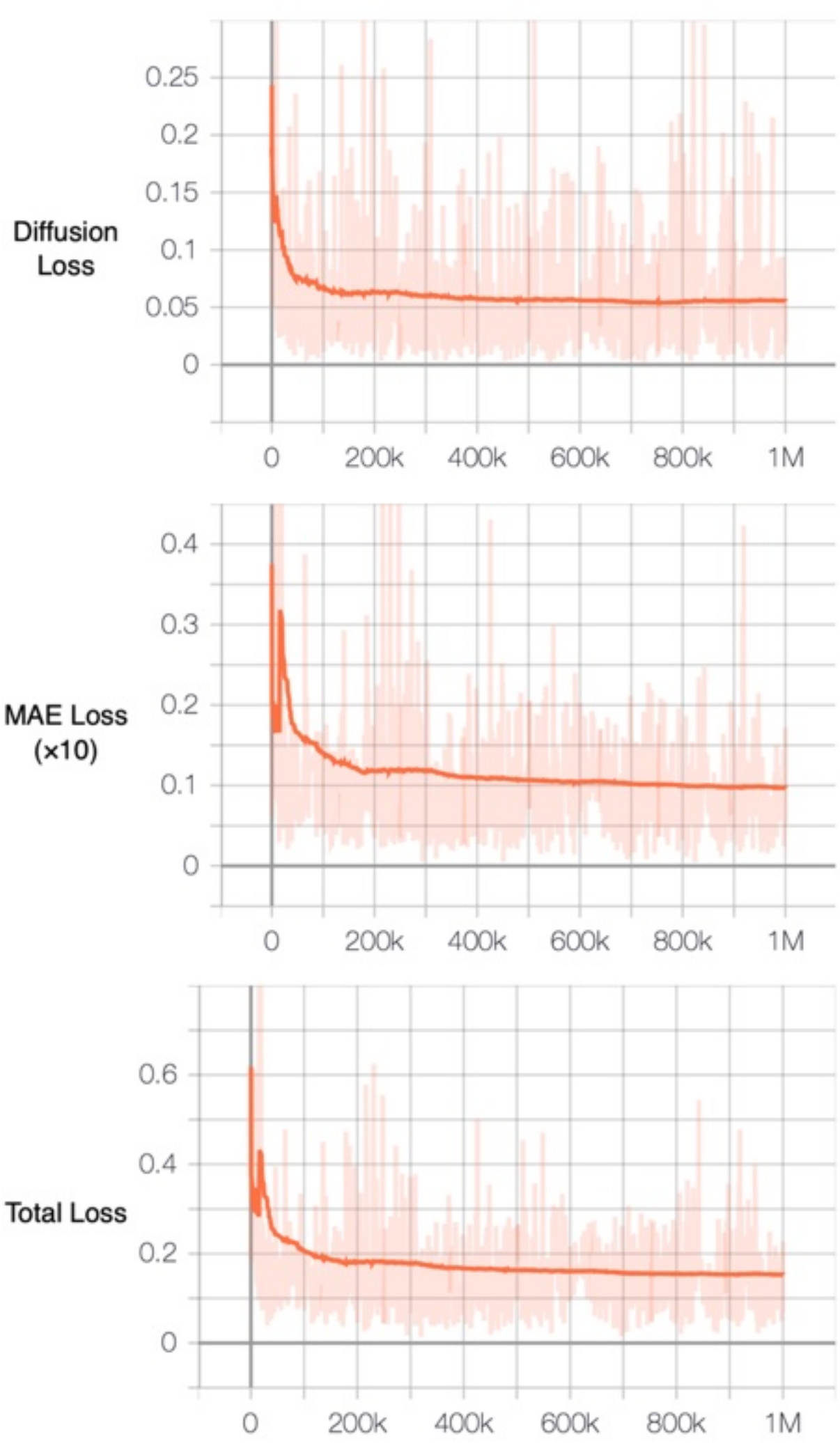

Figure 9: The loss curves of DDMS with respect to the training step. The diffusion and MAE losses are employed to train the satellite predictor and temporal diffusion part in DDMS,

respectively. The total loss is the combination of the diffusion loss and MAE loss. All the losses are steadily declining as the training step increases.

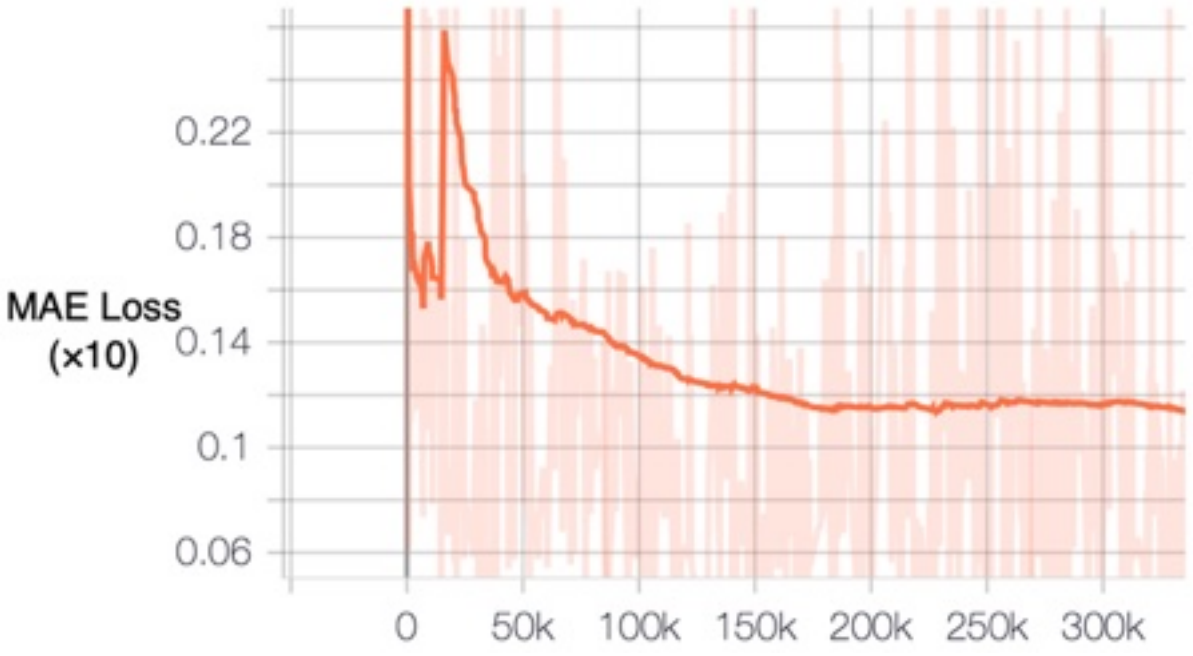

Figure 10: The loss curve of satellite predictor in DDMS with respect to the training step. In the early stage, the MAE loss dramatically increases during a short period and then steadily declines.

**2.2 Prediction Samples of DDMS and Its Components**

To visually compare the function of key components in DDMS, we give 4-hour prediction samples of DDMS and its components in Figures 11 - 23. In the proposed DDMS, the satellite predictor aims to make predictions with historical frames while the temporal diffusion part accounts for reconstructing the difference sequence. The final prediction is obtained by combining results produced by the satellite predictor and temporal diffusion part. By observing the samples, we have the following findings here. First, the frames produced by the satellite predictor become increasingly blurry as the lead time passes. This is reasonable since the satellite predictor tends to produce average results with the guidance of the MAE loss. Second, the temporal diffusion part nicely models the discrepancy between the prediction of the satellite predictor and the ground truth, which is vital for predicting some unpredictable motion trends of convection clouds such as dissipation and generation. Third, with the developed satellite predictor and temporal diffusion part, our DDMS delivers high-quality satellite nowcasting results both in terms of appearance details and motion trends. All the observations demonstrate the effectiveness of the proposed DDMS and its components.

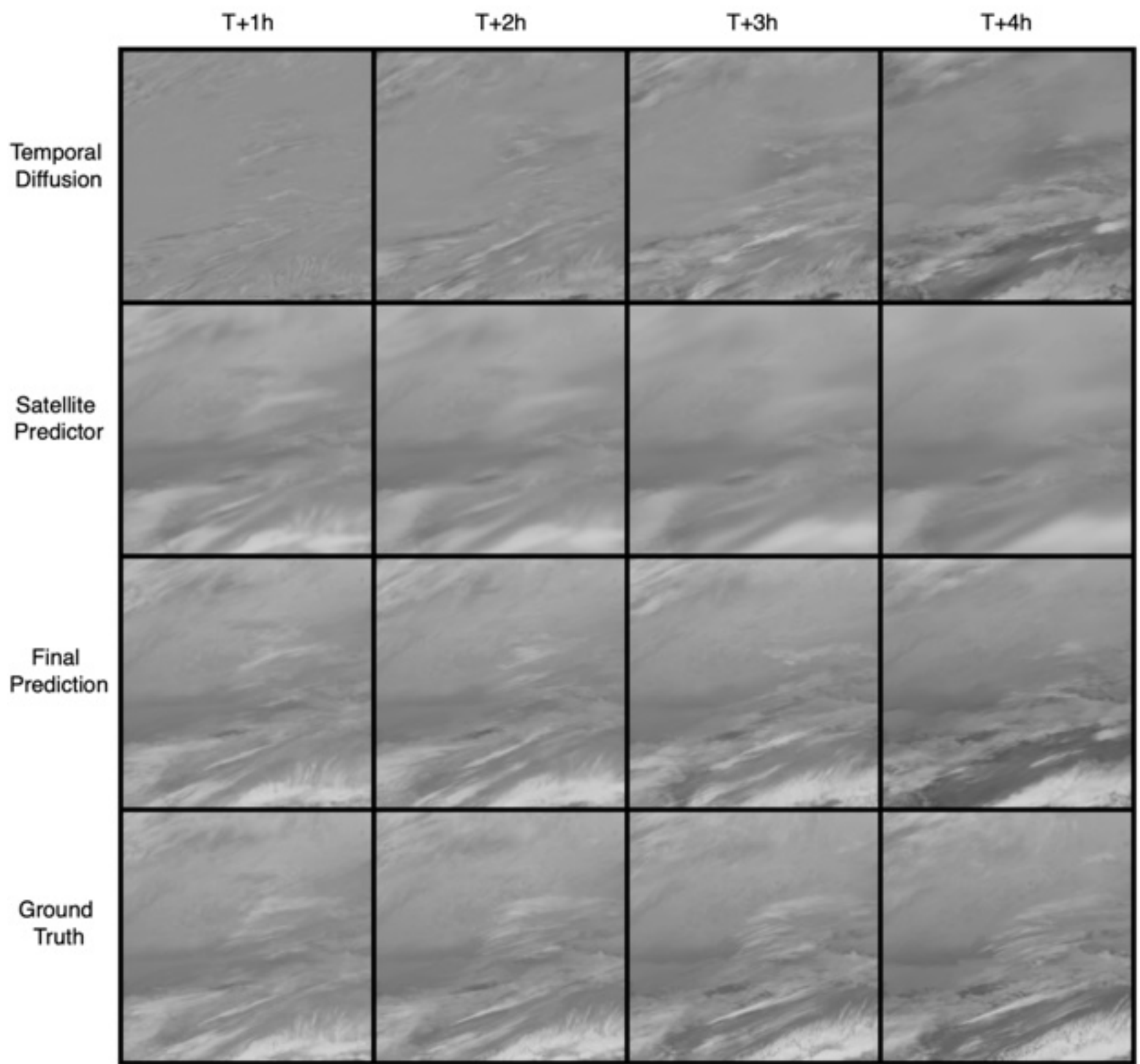

Figure 11: The 4-hour 256 × 256-size satellite nowcasting sample 1. Line 1 denotes the reconstructed difference sequence produced by the temporal diffusion part in DDMS. Line 2 denotes the satellite sequence predicted by the satellite predictor in DDMS. Line 3 denotes the final prediction results of DDMS. Line 4 represents the ground-truth sequence. In particular, the size of each frame is 256 × 256-size and every four frames are shown.

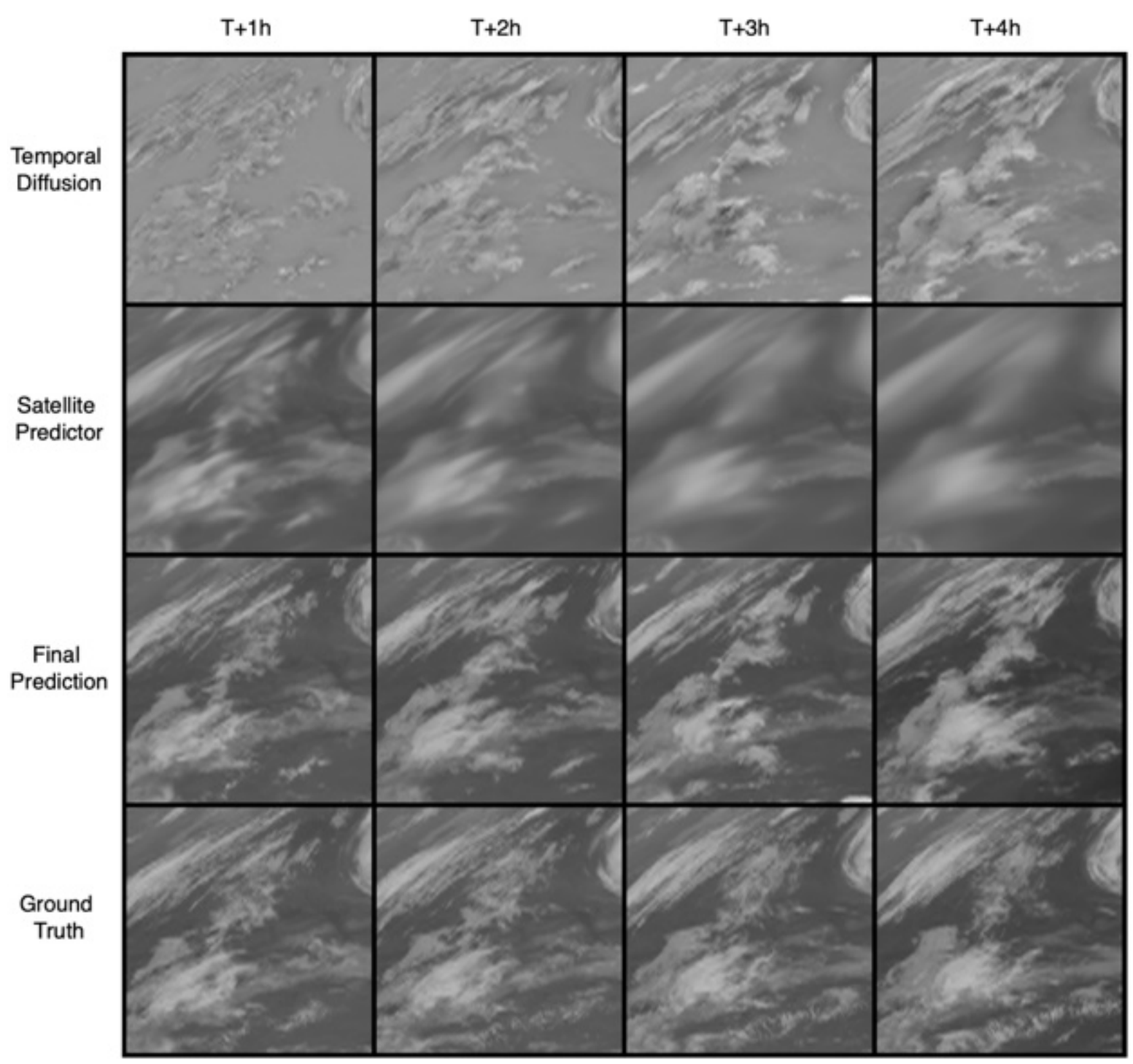

Figure 12: The 4-hour 256 × 256-size satellite nowcasting sample 2.

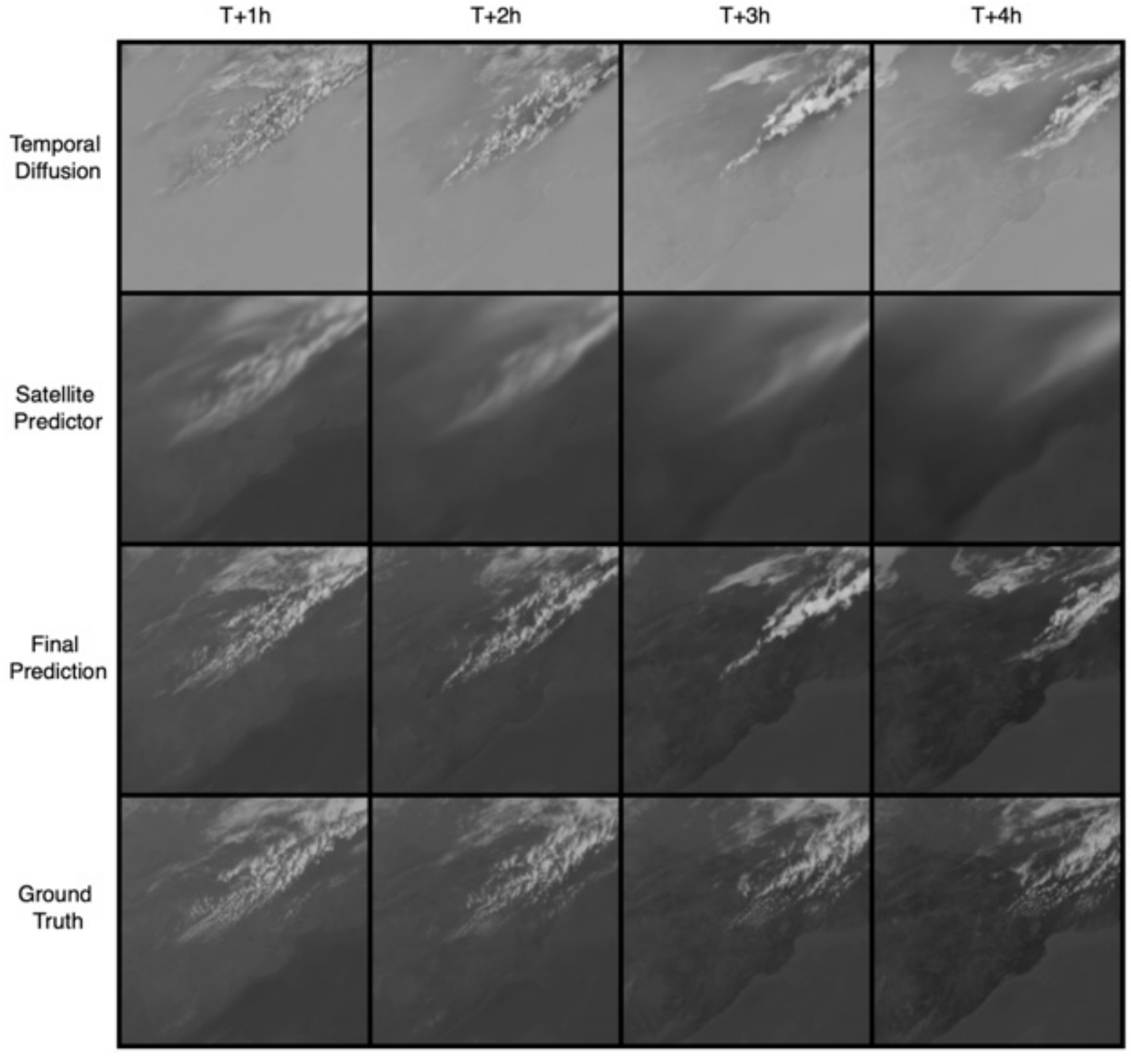

Figure 13: The 4-hour 256 × 256-size satellite nowcasting sample 3.

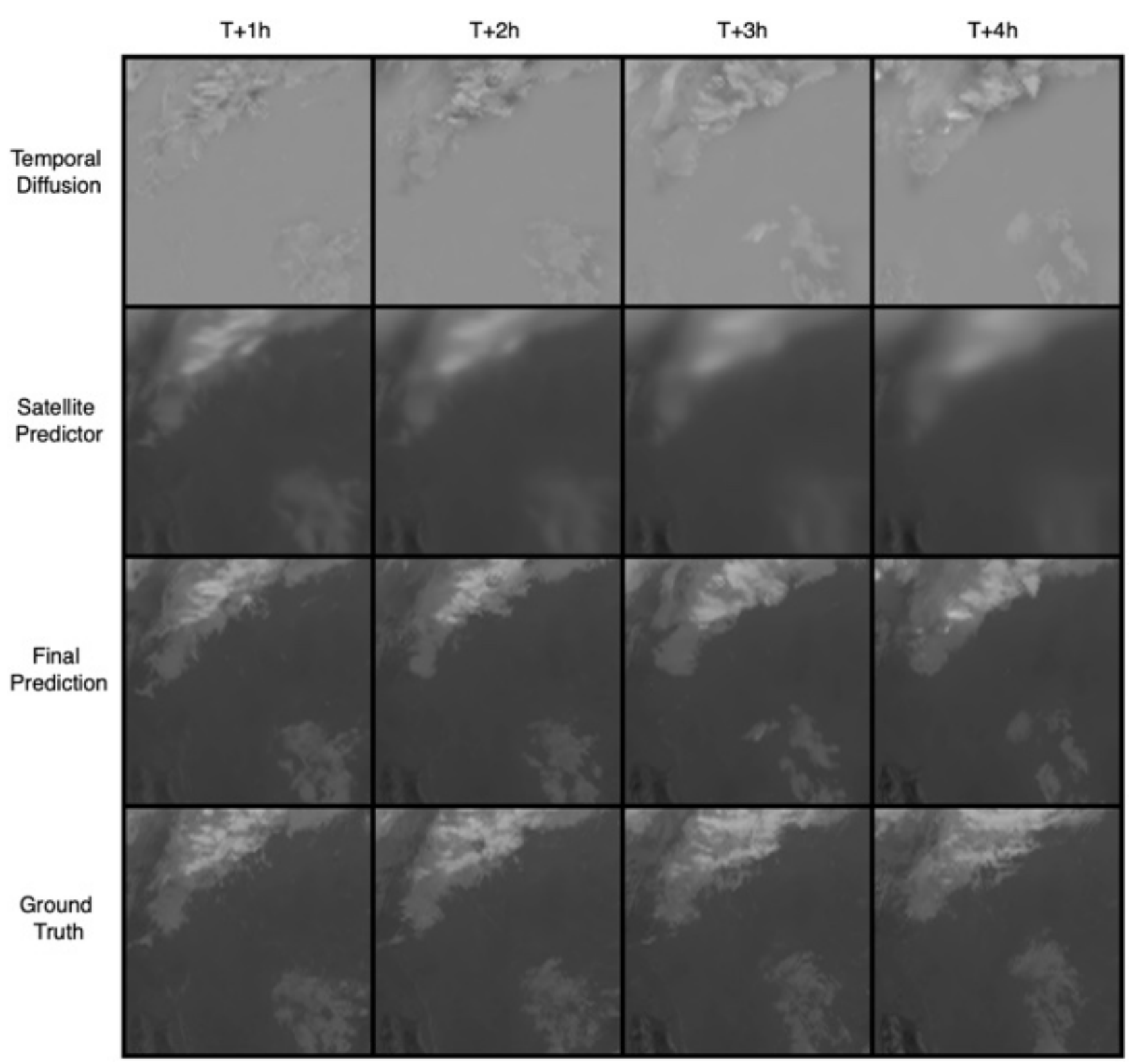

Figure 14: The 4-hour 256 × 256-size satellite nowcasting sample 4.

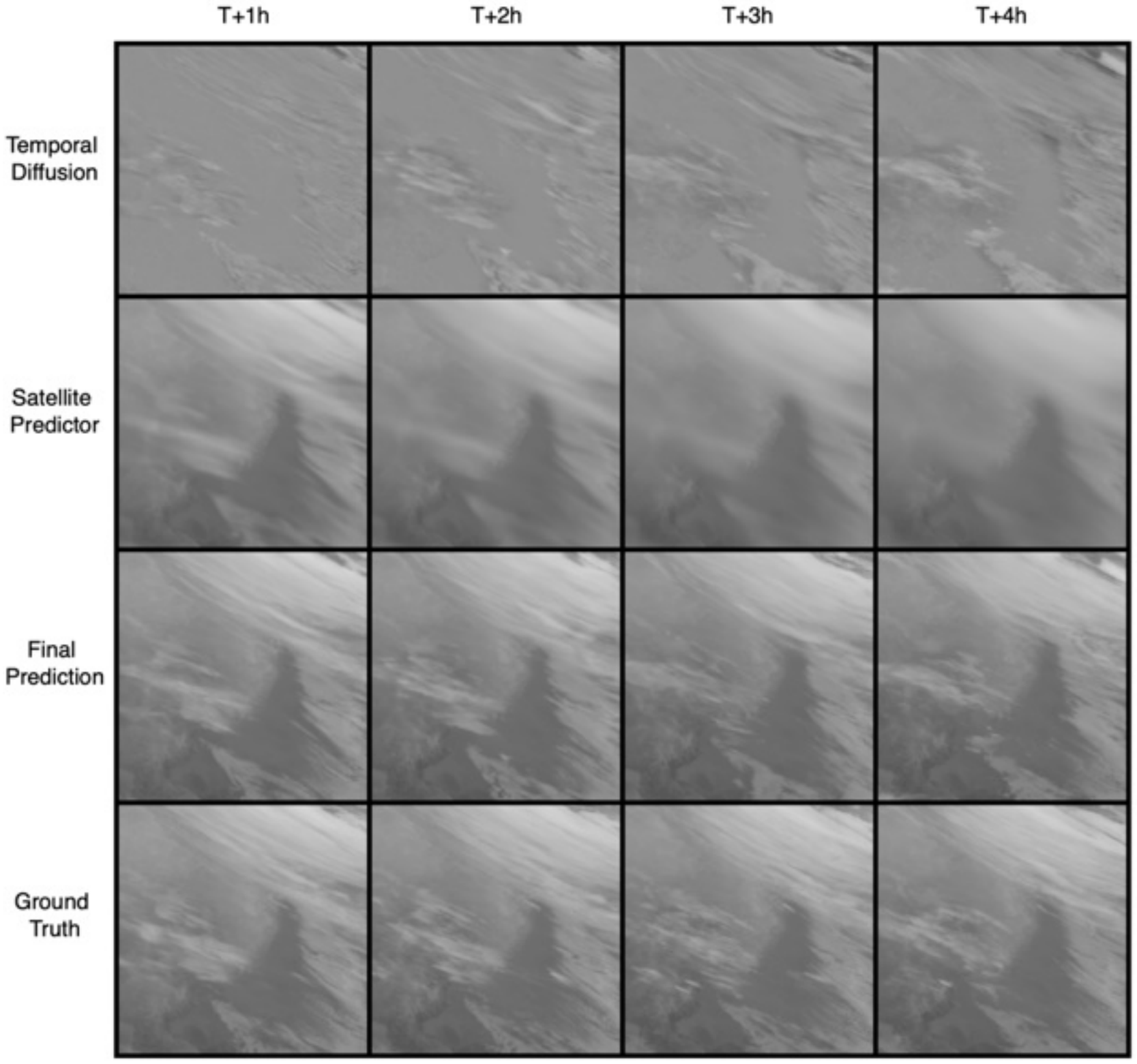

Figure 15: The 4-hour 256 × 256-size satellite nowcasting sample 5.

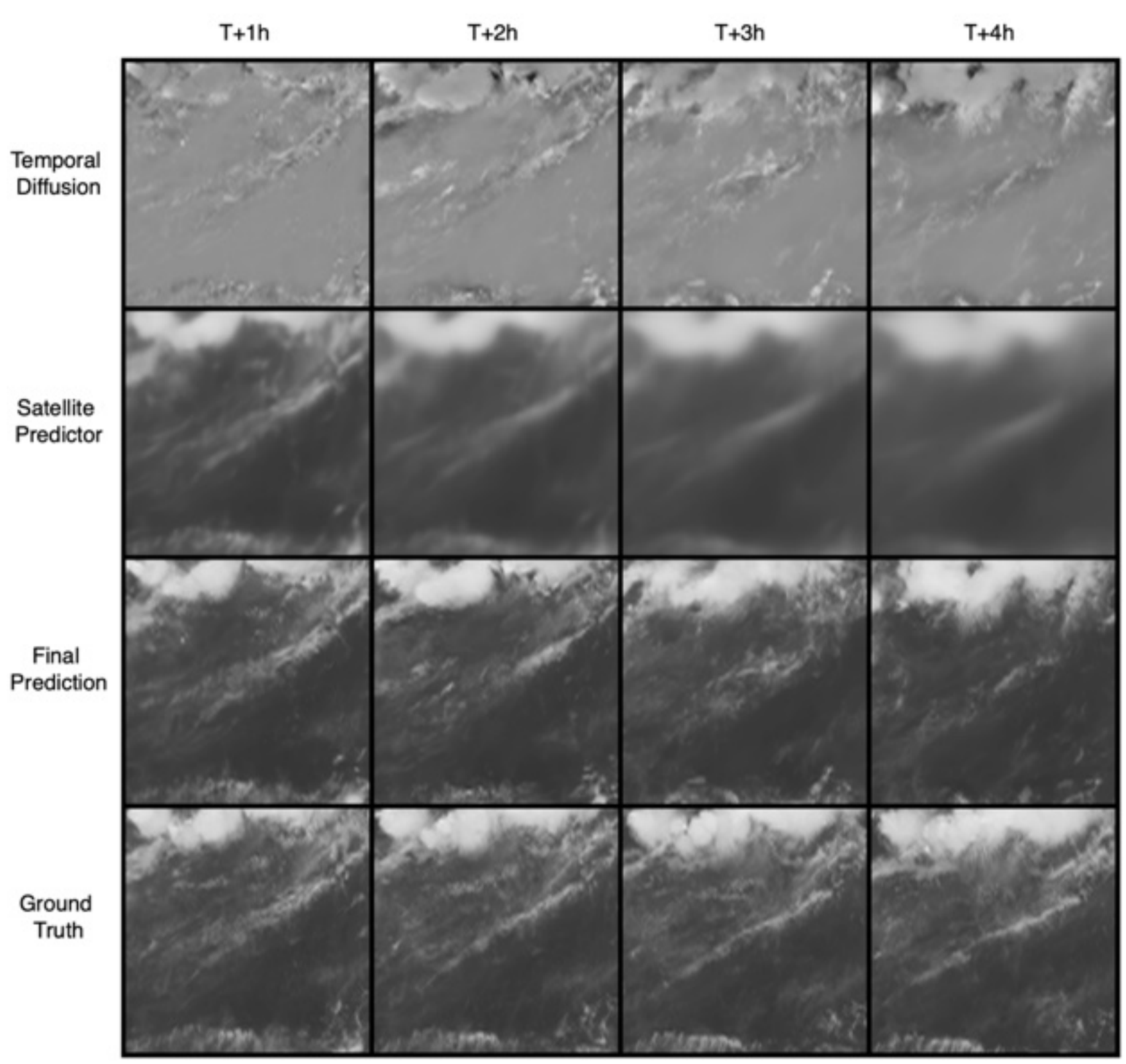

Figure 16: The 4-hour 256 × 256-size satellite nowcasting sample 6.

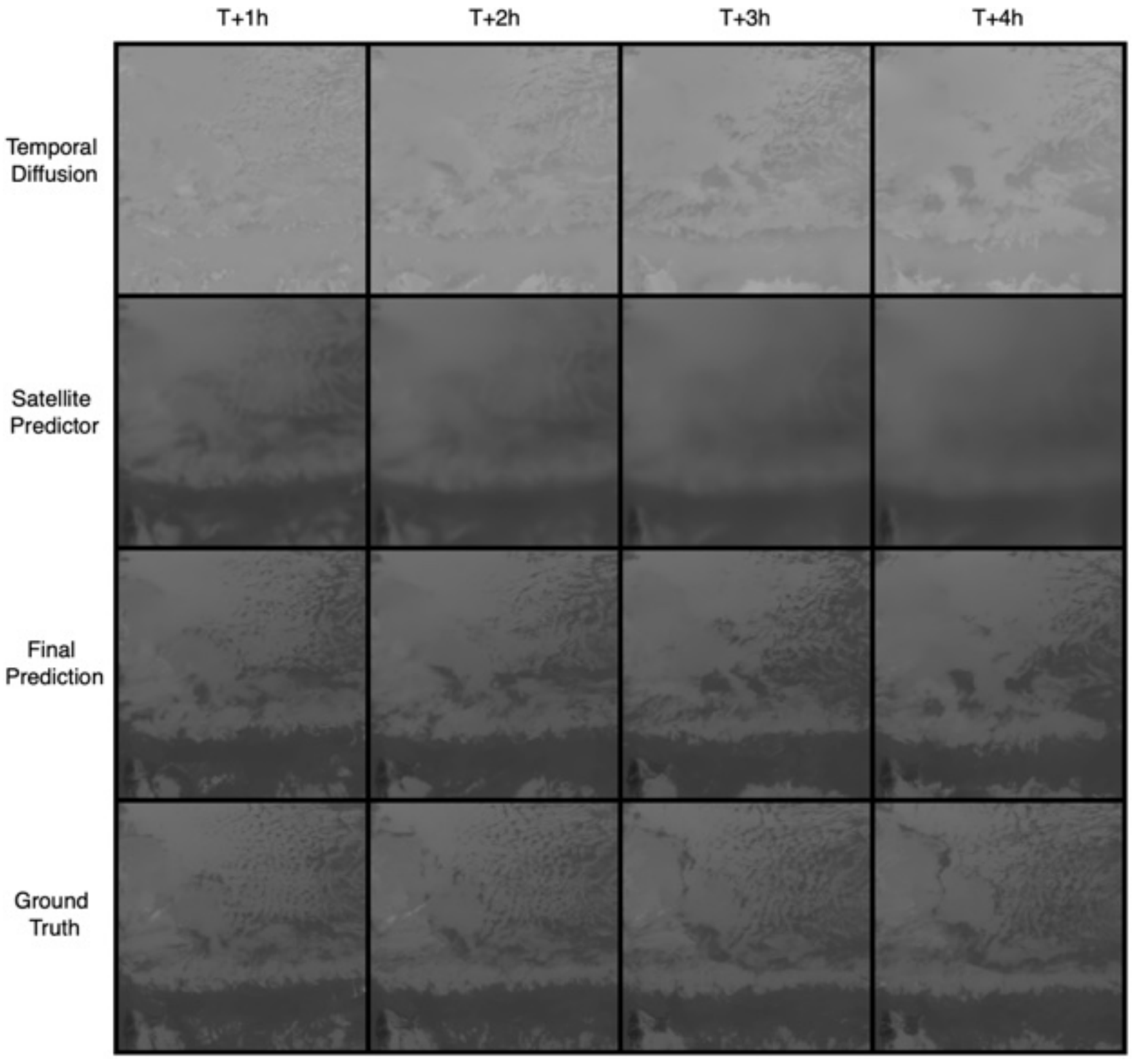

Figure 17: The 4-hour 256 × 256-size satellite nowcasting sample 7.

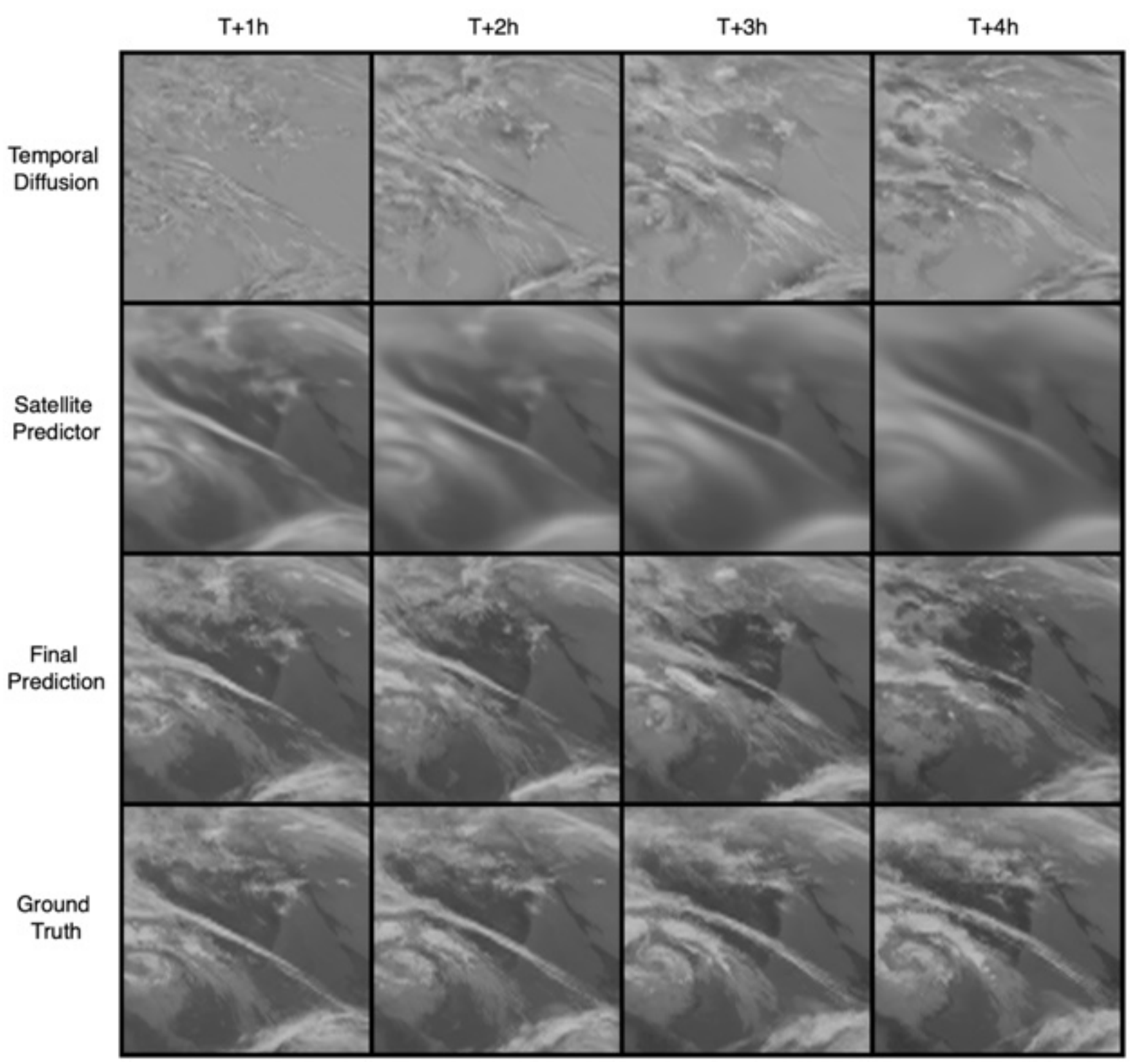

Figure 18: The 4-hour 256 × 256-size satellite nowcasting sample 8.

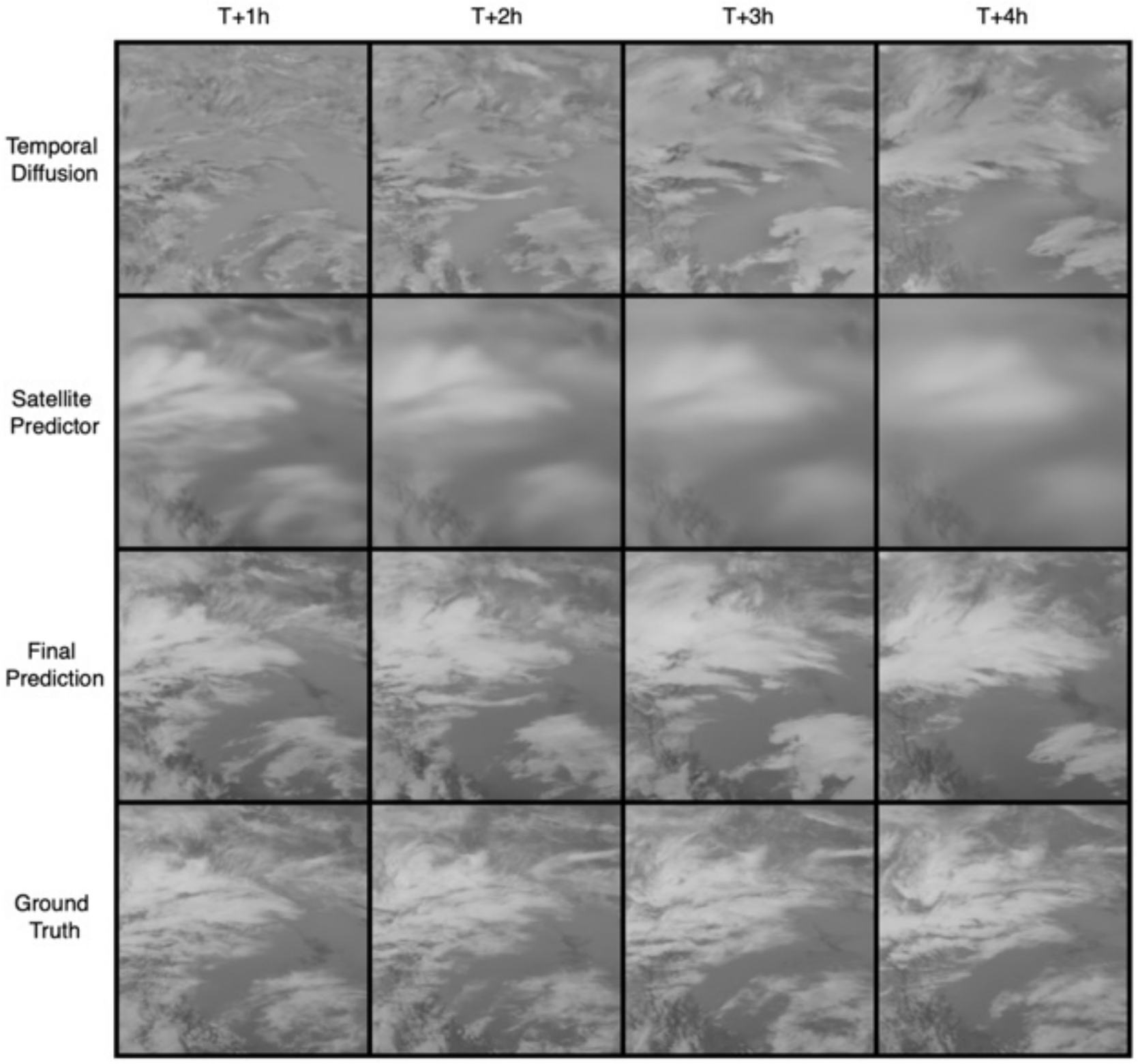

Figure 19: The 4-hour 256 × 256-size satellite nowcasting sample 9.

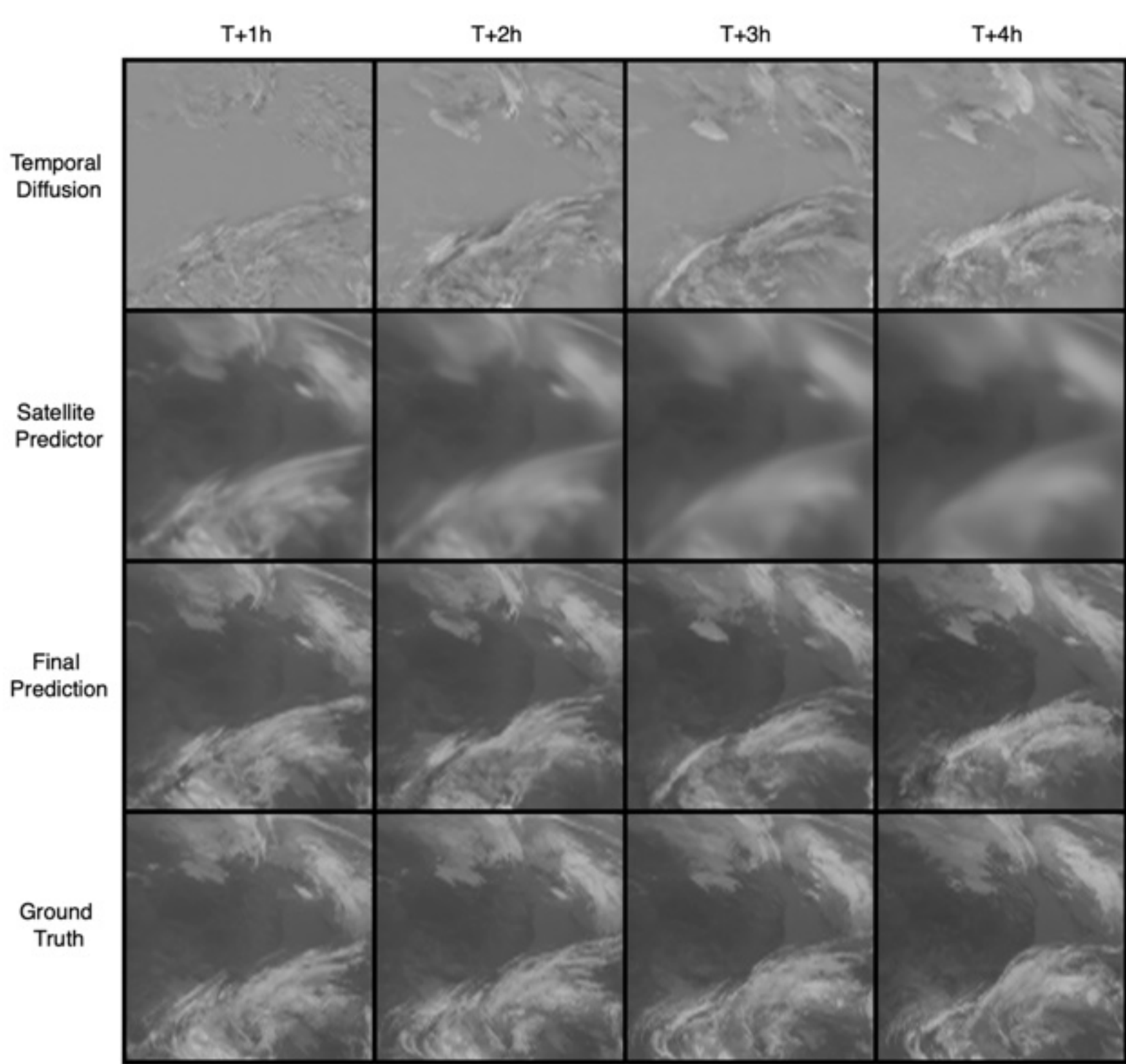

Figure 20: The 4-hour 256 × 256-size satellite nowcasting sample 10.

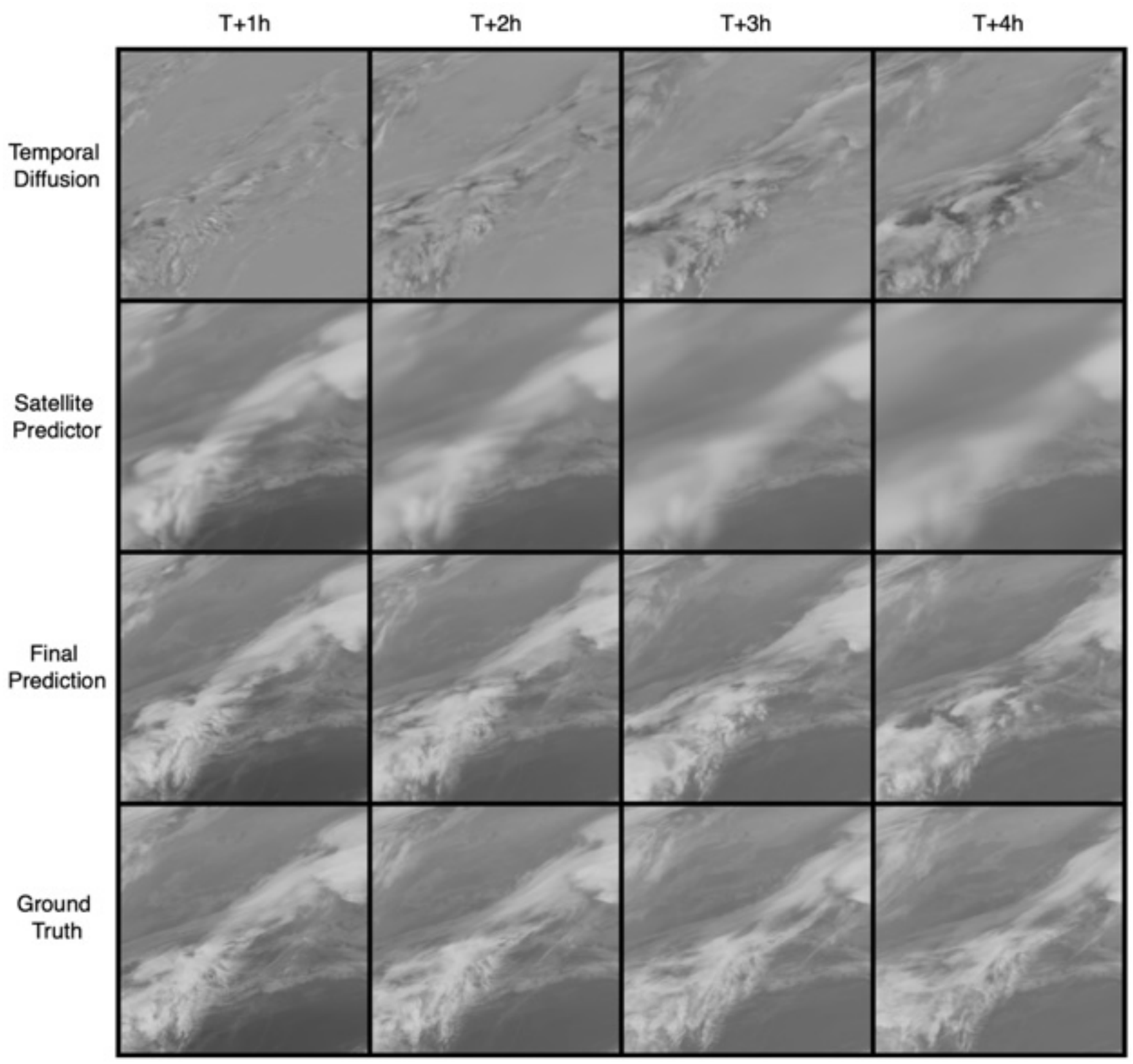

Figure 21: The 4-hour 256 × 256-size satellite nowcasting sample 11.

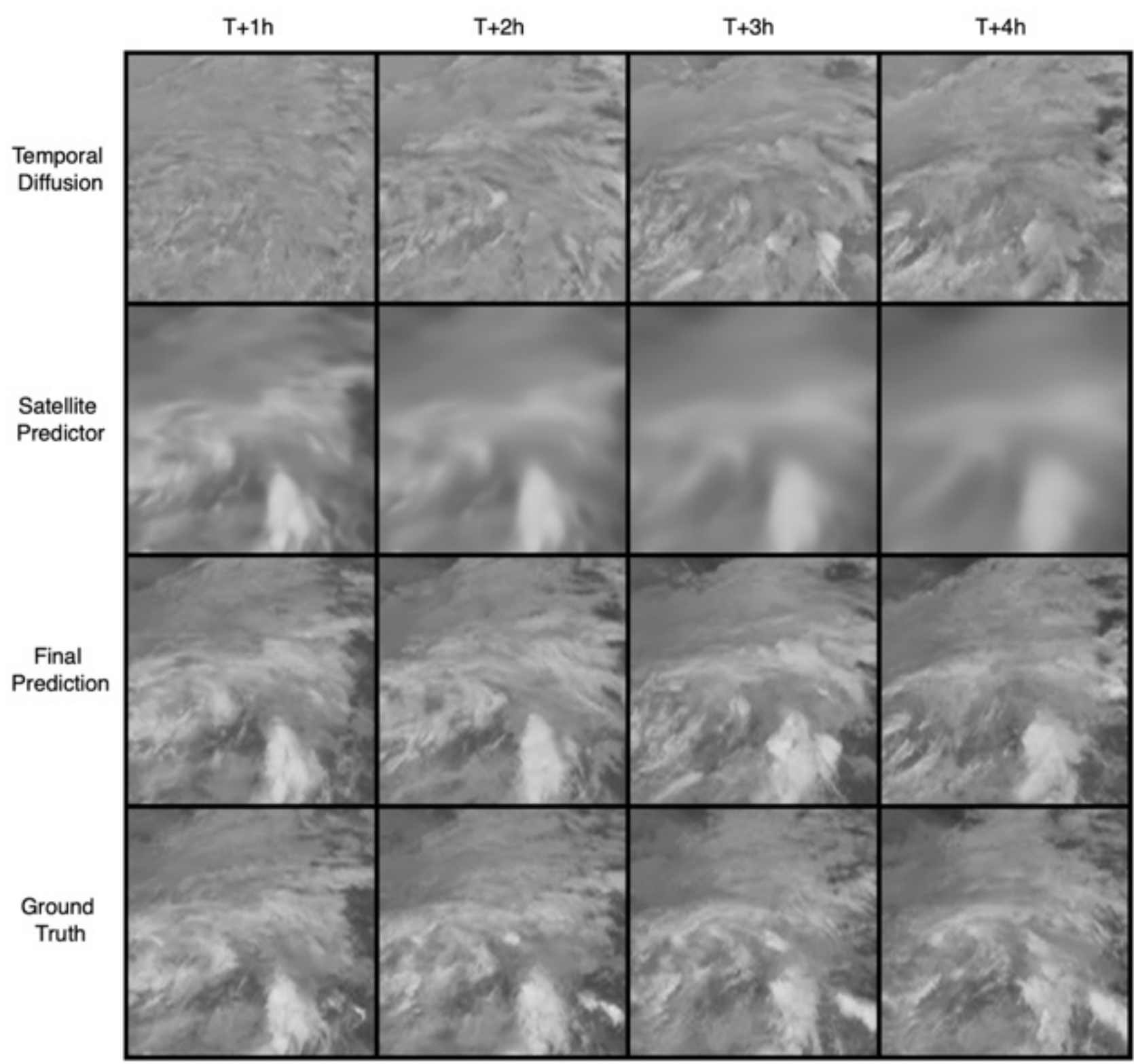

Figure 22: The 4-hour 256 × 256-size satellite nowcasting sample 12.

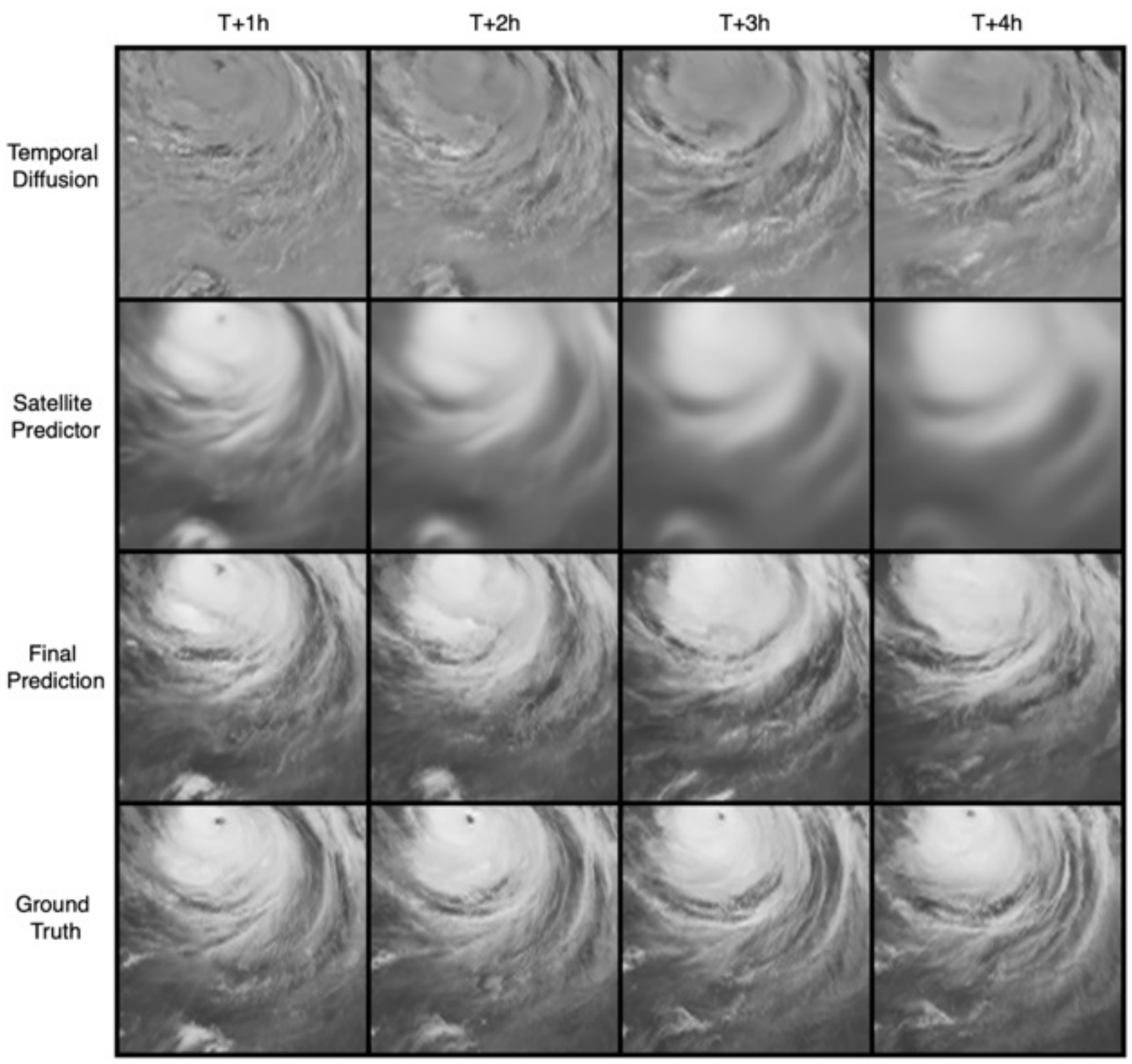

Figure 23: The 4-hour 256 × 256-size satellite nowcasting sample 13.

## 3. Scheduled Temporal Diffusion Weights

As shown in subsection 2.2, the designed temporal diffusion module can effectively model spatiotemporal evolution patterns of convective clouds. However, the prediction accuracy becomes harder to maintain since the uncertainty of dynamics in the atmosphere rapidly increases as the time step advances. To address this issue, we propose a corresponding scheduled temporal weights strategy to cooperate with the temporal diffusion module to guarantee prediction accuracy. In particular, detailed temporal weights are formally defined as follows:

$$i = 0,1,2,\dots,15,$$
$$\text{weights}_i = w \cdot 0.5 \cdot \left(1 + \cos\left(\frac{i\pi}{15}\right)\right) + (1 - w).$$

The $\text{weights}_i$ denotes the corresponding value of each prediction time step. The $w$ is set to 0.4. By applying the scheduled temporal weights to the reconstructed difference frame produced by the temporal diffusion module, DDMS can deliver better prediction results. Specifically, the quantitative nowcasting results at the 4km scale of the original DDMS and the variant with the scheduled weights are shown in Table 1. Except for the CSI, we utilize three commonly used metrics in weather nowcasting including false alarm rate (FAR), probability of detection (POD), and Heidke skill score (HSS) to comprehensively measure prediction accuracy. The FAR score denotes the false alarm rate, which is the lower the better. The POD score represents the recall rate, which is the higher the better. The HSS score measures the improvement over stochastic prediction, which is the higher the better. Intuitively, DDMS significantly promotes its prediction ability with the scheduled temporal weights. Across the eight months, DDMS w/ weights significantly outperforms DDMS on FAR, CSI, and HSS scores while achieving competitive performance on the POD score. The observation demonstrates the effectiveness of the scheduled temporal weights strategy.

Table 1: Quantitative convection nowcasting results of DDMS and its variant with scheduled weights. The bold denotes better performance.

| Year | Model | Month | POD ↑ | FAR ↓ | CSI ↑ | HSS ↑ |
|---|---|---|---|---|---|---|
| 2022 | DDMS | 202205 | **0.457** | 0.558 | 0.318 | 0.442 |
| | DDMS w/ weights | 202205 | 0.429 | **0.481** | **0.329** | **0.453** |
| | DDMS | 202206 | **0.518** | 0.471 | 0.381 | 0.513 |
| | DDMS w/ weights | 202206 | 0.506 | **0.398** | **0.399** | **0.534** |
| | DDMS | 202207 | **0.481** | 0.500 | 0.351 | 0.479 |
| | DDMS w/ weights | 202207 | 0.457 | **0.425** | **0.360** | **0.490** |
| | DDMS | 202208 | **0.505** | 0.474 | 0.372 | 0.503 |
| | DDMS w/ weights | 202308 | 0.490 | **0.411** | **0.385** | **0.518** |
| 2023 | DDMS | 202305 | **0.399** | 0.612 | 0.274 | 0.384 |
| | DDMS w/ weights | 202305 | 0.369 | **0.575** | **0.273** | **0.380** |
| | DDMS | 202306 | **0.514** | 0.485 | 0.374 | 0.504 |
| | DDMS w/ weights | 202306 | 0.499 | **0.453** | **0.380** | **0.511** |
| | DDMS | 202307 | **0.489** | 0.492 | 0.357 | 0.487 |
| | DDMS w/ weights | 202307 | 0.474 | **0.459** | **0.361** | **0.492** |
| | DDMS | 202308 | **0.487** | 0.500 | 0.353 | 0.481 |
| | DDMS w/ weights | 202308 | 0.473 | **0.466** | **0.358** | **0.488** |

## 4. Convective Clouds Label Principles

Due to the significant influence of geographical location and seasonal factors on the formation and presence of convective clouds, manually labeling convective clouds based on satellite brightness images, satellite visible images, and radar echo images is pretty difficult. For this, our team and experts in the National Satellite Meteorological Center have summarized basic label principles, which are described as follows:

1. The important convective clouds generally range in size from 2 km to 200 km, with a few reaching up to 300 km. If a large convective cloud of around 1000 km appears, it is composed of multiple convective clouds.
2. Independent convective clouds appear in clusters with clear boundaries, looking denser than ordinary clouds. Striped clouds are generally not convective clouds.
3. Clouds with gradually decreasing brightness temperature are most likely convective clouds.
4. If the cloud is large enough or the brightness temperature drops quickly enough, it is very likely to be a convective cloud.
5. The brightness temperature range is narrower for ordinary clouds compared to convective clouds.
6. In visible channels, the presence of bubbling upwards typically indicates convective clouds. When the cloud tops are smooth or obscured by regular clouds, judgments for convective clouds should integrate changes in brightness temperature images or radar data.
7. The brightness temperature decreases over a large cloud system, which is not composed of multiple convective clouds, generally indicates a widespread climate change, such as the arrival of cold air, rather than convective clouds formation.
8. The convective cloud characteristics over the sea are quite distinct, densely clustered formations are typically convective clouds.
9. Generally, convective clouds do not form in northern China during winter. Without clear convective cloud characteristics, it is not typically considered to be a convective cloud.

10. In northeast China, convective clouds often accompany the appearance of vortex cloud systems, with convective clouds typically at the periphery of the vortex center.
11. During summer, small convective clouds appear almost daily over the Qinghai-Tibet Plateau, blending within extensive cloud formations, with lower accuracy in identification.
12. Strong radar echoes indicate a high probability of precipitation, suggesting that convective clouds are likely present in the area.